%% file: resub.tex
\documentclass{llncs}
\pdfoutput=1   
\usepackage[T1]{fontenc}

\usepackage{graphicx}

\usepackage{amsmath,amssymb}
\usepackage{booktabs}
\usepackage{longtable}
\usepackage{array}
\usepackage{pifont}
\usepackage{pdflscape}   

\usepackage{tikz}
\usetikzlibrary{arrows.meta,positioning,shapes.geometric}

\usepackage[hidelinks]{hyperref}

\providecommand{\doi}[1]{\textsc{doi}:\ \href{https://doi.org/#1}{#1}}

\newcommand{\cmark}{\ding{51}} 
\newcommand{\xmark}{\ding{55}} 

\begin{document}
\title{Real-Time QP Solvers: A Review and Practical Guide for Legged Robots}

\author{Van-Nam Dinh\inst{1}}
\institute{Institute of Science and Technology, Vinh University, Vinh, Nghe An, Viet Nam\\
\email{namdv@vinhuni.edu.vn}
}

\maketitle
\pagestyle{plain}

%
\begin{abstract}
Model Predictive Control (MPC) and Whole-Body Control (WBC) are both posed as
quadratic programs (QPs) and solved under hard timing budgets on embedded
hardware. We review the solver families used for them: active-set,
interior-point, operator-splitting and augmented-Lagrangian/proximal methods,
together with solver code generation. We then consolidate the public
benchmarking evidence, which aggregates solve times on problem sets that
cannot be re-run and reports speed more thoroughly than the accuracy attained
at default settings. Our open benchmark is built on the two QP templates of a
legged stack: single-rigid-body-dynamics MPC, sparse and condensed, and
whole-body inverse dynamics. Fourteen solvers are measured for solve time,
attained residual, warm-start benefit and infeasibility behavior, in Python
and, for three that have no binding, in C++. Keeping the optimal-control
sparsity is $4.5\times$ faster than condensing at a horizon of $40$ and
$1.7\times$ slower at a horizon of $10$, so the familiar advice to exploit
sparsity is horizon-dependent. At shipped defaults one solver takes $4.56$
seconds where its own MPC preset takes $20$\,ms, and another declares a
feasible condensed MPC problem primal infeasible after one iteration, on two
platforms, against a two-line analytic feasibility certificate.
Operator-splitting solvers stop many orders of magnitude short of
interior-point and active-set solvers; at a matched tolerance the
speed--accuracy trade-off reappears at a factor of $2.2$ to $8.7$. Most
reported warm-start benefit is setup reuse: $24.8\times$ from keeping a dual
active-set solver's object alive against $1.22\times$ from its warm start, the
one unambiguous gain being $1.56\times$ for a proximal method, which ships
with it off. A third solver's warm start is worth up to $5.6\times$ and is
hidden entirely by its default termination-check interval. A GPU round trip
for the whole-body problem exceeds a single-core CPU solve, and the device
needs tens of simultaneous problems to break even, so the boundary of useful
GPU acceleration falls between the two templates rather than around them. We
give a solver-selection procedure and release the generator and harness, so
every number here can be regenerated.
\keywords{Quadratic Programming \and Optimization \and Real-Time Control \and Legged Robots \and QP Solvers \and Benchmarking \and GPU Acceleration \and Reproducibility}
\end{abstract}

\section{Introduction}
Real-time optimization forms the backbone of advanced legged robot control~\cite{Humanoid_rev}, including state estimation~\cite{van2024fusion,namlearning}, motion planning~\cite{Humanoid_rev}, and dynamic control~\cite{dicarlo2018,wensing2023}.
Quadratic programming (QP)~\cite{ProxQP} enables the formulation of control objectives and constraints in a convex, computationally efficient, and theoretically grounded framework, and it is widely adopted in legged and humanoid control stacks under stringent latency and energy constraints~\cite{verschueren2018thesis,acados}.

In a standard legged robot control architecture~\cite{dicarlo2018}, a \textit{Model Predictive Controller} (MPC) computes contact forces or center-of-mass trajectories over a finite horizon, often yielding sparse optimal-control-structured QPs~\cite{OCS2,acados}. In contrast, a \textit{Whole-Body Controller} (WBC) converts these high-level commands into joint torques or velocities subject to actuation, friction, and contact constraints, typically resulting in small-to-medium dense QPs solved at high rates (100--1000\,Hz)~\cite{wensing2023}. The two layers run at different rates, and that separation is what makes solver selection template-specific: the MPC layer is re-solved at tens of hertz over a horizon of a few hundred milliseconds, while the whole-body layer runs at the servo rate~\cite{kim2019,grandia2023}. In both cases, solver speed, memory resources, numerical robustness, and warm-starting determine closed-loop feasibility.

Recent benchmarking efforts highlight that solver selection is strongly scenario- and structure-dependent~\cite{stark2025benchmarking,qpbenchmark,qpmad_benchmark}. For sparse MPC QPs in OCP form, structure-exploiting interior-point solvers such as \texttt{HPIPM}~\cite{HPIPM} are often a top choice; for large generic sparse QPs or when moderate accuracy suffices, operator-splitting solvers such as \texttt{OSQP}~\cite{OSQP} are attractive due to warm-starting and predictable iteration costs. For dense WBC QPs, active-set solvers such as \texttt{qpOASES}~\cite{qpOASES}, \texttt{Eiquadprog}~\cite{Eiquadprog}, and the lightweight \texttt{qpmad}~\cite{qpmad} remain strong baselines.
Beyond these widely used tools, modern solver options include embedded-focused dual active-set solvers such as \texttt{DAQP}~\cite{daqp}, proximal interior-point solvers designed for ill-conditioned sparse QPs such as \texttt{PIQP}~\cite{piqp}, and sparse primal-dual IPMs positioned for robotics such as \texttt{qpSWIFT}~\cite{qpSWIFT}. In industrial deployment, code-generated solvers (e.g., \texttt{FORCESPRO}~\cite{forcespro}, \texttt{CVXGEN}~\cite{cvxgen}, \texttt{CVXPYgen}~\cite{cvxpygen}) can provide highly optimized C implementations for fixed-structure problem families. Finally, when conic constraints or certificates are required, conic IPMs such as \texttt{Clarabel}~\cite{clarabel} provide a general convex optimization route that also covers QPs.

Most prior comparisons emphasize solve time and numerical accuracy~\cite{qpbenchmark}, but for battery-powered robots, the energy cost of optimization is often equally critical.
Recent work therefore advocates energy-normalized metrics such as Solve Frequency per Watt (SFPW), enabling fairer comparisons across heterogeneous CPU architectures (e.g., ARM versus x86)~\cite{stark2025benchmarking}.
We report the SFPW evidence of~\cite{stark2025benchmarking} where it bears on solver--hardware interactions, but we measure no energy ourselves (Sect.~\ref{sec:limits}).

Four prior reviews overlap with this one and bound its contribution.
Wensing et al.~\cite{wensing2023} and Carpentier and Wieber~\cite{carpentier2021}
survey optimization-based legged control at the level of formulations rather
than solvers; Katayama et al.~\cite{katayama2023} survey the model and
algorithm choices of legged and humanoid MPC, again without comparative solver
measurements; and Kouzoupis et al.~\cite{kouzoupis2018} review the QP
algorithm families of Sect.~\ref{sec:qp} for nonlinear MPC generally, without
the legged-specific instances or the default-setting and interface effects
measured here.

Building on this context, this paper makes four contributions.
First, we review the QP solver families relevant to real-time control in
robotics and summarize representative open-source and commercial solvers
(Sect.~\ref{sec:qp}).
Second, we consolidate publicly available benchmarking evidence from
\texttt{qpbenchmark}~\cite{qpbenchmark},
\texttt{qpmad\_benchmark}~\cite{qpmad_benchmark} and the quadrupedal study
of~\cite{stark2025benchmarking} (Sect.~\ref{sec:public}).
Third, because that published evidence reports aggregate timings on
problem sets that are not released in a directly re-runnable form, and
reports speed far more thoroughly than the accuracy attained, we add our
own open benchmark, built directly on the two QP templates of a legged-robot
stack, Eqs.~\eqref{eq:mpc} and~\eqref{eq:wbc}. It covers fourteen solvers in
both the sparse and the condensed MPC formulation, measured through a Python
interface and, for the three solvers that have no binding, natively in C++,
and it reports solution accuracy, warm-start behavior and response to
infeasibility as well as speed (Sect.~\ref{sec:own-bench}).
Fourth, we review GPU-accelerated QP and MPC and measure the fixed cost of
using a GPU for problems of this size, which locates the boundary of useful
GPU acceleration inside a legged-robot stack and not around it
(Sect.~\ref{sec:gpu}).
The generator and both measurement harnesses are released with the paper so
that every number below can be regenerated with a single command.

\section{Quadratic Programming in Legged-Robot Control}
\label{sec:qp}
We use $n$ for the number of decision variables, $n_e$ for the number of
equality constraints, and $n_i$ for the number of inequality constraints.
Because this paper moves between the optimization and the robotics
conventions, two symbols are renamed from their most common form: the NLP
objective is written $\phi(\cdot)$, leaving $f$ free for the contact forces of
Eq.~\eqref{eq:wbc}, and the inequality upper bound of the standard QP form is
written $d$, leaving $u_k$ free for the control input of Eq.~\eqref{eq:mpc}.

In Sequential Quadratic Programming (SQP)~\cite{acados}, a nonlinear program (NLP) is solved iteratively by approximating it locally with a quadratic program (QP). At each iteration, the nonlinear objective function is replaced by a second-order Taylor expansion of the Lagrangian, while equality and inequality constraints are linearized. Specifically, consider the NLP
\[
\min_{x \in \mathbb{R}^n} \phi(x) \quad \text{s.t.} \quad h(x) = 0,\; g(x) \le 0,
\]
with $h:\mathbb{R}^n\to\mathbb{R}^{n_e}$ and $g:\mathbb{R}^n\to\mathbb{R}^{n_i}$. The SQP subproblem around the current iterate $x_j$ is
\[
\begin{aligned}
\min_{\Delta x \in \mathbb{R}^n} \quad &
\frac{1}{2} \Delta x^\top \tilde{H}_j \Delta x + \nabla \phi(x_j)^\top \Delta x \\
\text{s.t.} \quad &
J_h(x_j)\, \Delta x + h(x_j) = 0,\\
& J_g(x_j)\, \Delta x + g(x_j) \le 0,
\end{aligned}
\]
where $J_h(x_j)=\frac{\partial h}{\partial x}(x_j)\in\mathbb{R}^{n_e\times n}$ and $J_g(x_j)=\frac{\partial g}{\partial x}(x_j)\in\mathbb{R}^{n_i\times n}$ are Jacobians, and $\tilde{H}_j$ approximates the Hessian of the Lagrangian. The iterate index $j$ is distinct from the stage index $k$ used from Eq.~\eqref{eq:mpc} onwards.

This subproblem is a convex QP when $\tilde{H}_j$ is positive semi-definite and fits the standard convex QP form~\cite{ProxQP}:
\begin{equation}
\label{eq:qp}
\begin{aligned}
\min_{x \in \mathbb{R}^n} \quad & \frac{1}{2} x^\top H x + q^\top x \\
\text{s.t.} \quad & A x = b, \\
                  & C x \le d,
\end{aligned}
\end{equation}
by identifying $H=\tilde{H}_j$, $q=\nabla \phi(x_j)$, $A=J_h(x_j)$, $b=-h(x_j)$, $C=J_g(x_j)$, and $d=-g(x_j)$, with decision variable $x \equiv \Delta x$.

In real-time deployment this loop is almost never run to convergence. Under
the real-time iteration scheme~\cite{diehl2005,gros2020} one QP is solved per
sampling instant and its solution applied immediately, the next linearization
taking over where the previous one stopped; \texttt{acados}~\cite{acados},
which dispatches to the solvers of Table~\ref{tab:qp-solvers} instead of
being one itself, operates this way, as does the quadrupedal study
of~\cite{stark2025benchmarking}. This is what the \emph{Early} column of
Table~\ref{tab:qp-solvers} is for: the figure of merit is the control
performance obtained from one inexact solve inside the sampling period, not
the time to a tight tolerance.

For problems with only equality constraints (no $C x \le d$), and provided
$A$ has full row rank and $H$ is positive definite on $\ker A$, the unique
solution is obtained from the Karush--Kuhn--Tucker (KKT) system:
\begin{equation}
\begin{bmatrix}
H & A^\top \\
A & 0
\end{bmatrix}
\begin{bmatrix}
x^* \\
\lambda^*
\end{bmatrix} =
\begin{bmatrix}
-q \\
b
\end{bmatrix},
\end{equation}
where $\lambda^*\in\mathbb{R}^{n_e}$ are the Lagrange multipliers for the equality constraints.

\subsection{The two QP families of a legged-robot control stack}
\label{sec:two-families}
The QPs that a legged robot solves online are not drawn from a generic
distribution: they come from two recurring templates whose structure differs
sharply, and that difference is what makes solver selection non-trivial.

\paragraph{(a) Centroidal MPC.}
Convex MPC over single-rigid-body dynamics (SRBD)~\cite{dicarlo2018} predicts
the body trajectory under contact forces. The state stacks orientation,
position, angular and linear velocity and a constant gravity entry,
\[
x_k=\bigl[\theta^\top,\; p^\top,\; \omega^\top,\; \dot{p}^\top,\;
\gamma\bigr]^\top \in\mathbb{R}^{n_x},\qquad n_x=13,
\]
with $\theta,p,\omega,\dot{p}\in\mathbb{R}^3$ and $\gamma\in\mathbb{R}$ the
constant gravitational acceleration; the input
$u_k\in\mathbb{R}^{n_u}$, $n_u=3n_c$, stacks the $n_c$ contact forces. The
body dynamics are affine in the state; carrying $\gamma$ as a thirteenth
entry absorbs the constant, so the discretized dynamics take the linear form
$x_{k+1}=A_k x_k + B_k u_k$ and the QP over a horizon $N$, given the measured
initial state $x_0$, reads
\begin{equation}
\label{eq:mpc}
\begin{aligned}
\min_{x_{1:N},\,u_{0:N-1}} \quad
& \tfrac{1}{2}\sum_{k=1}^{N} \|x_k-x_k^{\mathrm{ref}}\|_{Q}^2
+ \tfrac{1}{2}\sum_{k=0}^{N-1} \|u_k\|_{R}^2 \\
\text{s.t.}\quad
& x_{k+1}=A_k x_k + B_k u_k, \qquad k=0,\dots,N-1,\\
& C_{\mathrm{fric}}\, u_k \le h_{\mathrm{fric}}, \qquad k=0,\dots,N-1,
\end{aligned}
\end{equation}
where $\|w\|_Q^2:=w^\top Q w$, $Q\succeq0$, $R\succ0$, and
$C_{\mathrm{fric}}$ collects the linearized friction pyramid and the
normal-force bounds $f_z\in[\underline{f}_z,\overline{f}_z]$ of every stance
foot. Equation~\eqref{eq:mpc} can be passed to a solver in two very different
ways. In the \emph{sparse}, or OCP, form the states remain decision variables
and the dynamics become equality constraints: the problem has
$N(n_x+n_u)$ variables but a block-banded, extremely sparse KKT matrix. In the
\emph{condensed} form the states and inputs are stacked as
$\mathbf{X}=[x_1^\top,\dots,x_N^\top]^\top$ and
$\mathbf{U}=[u_0^\top,\dots,u_{N-1}^\top]^\top$, and the states eliminated
through $\mathbf{X} = S_x x_0 + S_u \mathbf{U}$ with
$S_x\in\mathbb{R}^{Nn_x\times n_x}$ and
$S_u\in\mathbb{R}^{Nn_x\times Nn_u}$, leaving only $N n_u$ variables but a
\emph{fully dense} Hessian $S_u^\top \bar{Q} S_u + \bar{R}$, where
$\bar{Q}=I_N\otimes Q$ and $\bar{R}=I_N\otimes R$. Condensing shrinks the
problem and destroys its sparsity at the same time, and as
Sect.~\ref{sec:own-bench} shows the change of form is enough to change which
solvers succeed. We deliberately draw no conclusion here from the two forms'
Hessian condition numbers: in this generator both are set by regularizers we
chose rather than by the dynamics. The condensed form's smallest eigenvalue
is exactly the input weight $R=10^{-5}$, and the sparse form's conditioning is
exactly $\lambda_{\max}(Q)$ divided by the ridge of Sect.~\ref{sec:protocol},
needed only because the gravity state carries no tracking weight. Comparing
them would compare our own knobs.

\paragraph{(b) Whole-body inverse dynamics.}
The WBC layer converts the MPC command into joint torques. With generalized
accelerations $\dot{v}\in\mathbb{R}^{n_v}$ and point-contact forces
$f\in\mathbb{R}^{3n_c}$ as decision variables, a standard stance-phase QP is
\begin{equation}
\label{eq:wbc}
\begin{aligned}
\min_{\dot{v},\,f} \quad
& \tfrac{1}{2}\|\dot{v}-\dot{v}^{\mathrm{des}}\|_{W}^2
+ \tfrac{1}{2}\varepsilon\|f\|_2^2 \\
\text{s.t.}\quad
& M_u\dot{v} + \eta_u = J_{c,u}^\top f, \\
& J_c\dot{v} + \dot{J}_c v = 0, \\
& C_{\mathrm{fric}} f \le h_{\mathrm{fric}}, \qquad
|\tau(\dot{v},f)| \le \bar{\tau},
\end{aligned}
\end{equation}
Here $M$ is the mass matrix, $\eta$ collects the Coriolis, centrifugal and
gravitational terms, $J_c\in\mathbb{R}^{3n_c\times n_v}$ is the stacked
contact Jacobian, and all of these are evaluated at the current configuration
and velocity. The subscript $u$ selects the six unactuated floating-base rows of
$M$ and $\eta$ and the corresponding six \emph{columns} of $J_c$,
equivalently the six unactuated rows of $J_c^\top$, so that
$M_u\in\mathbb{R}^{6\times n_v}$, $\eta_u\in\mathbb{R}^{6}$ and
$J_{c,u}\in\mathbb{R}^{3n_c\times6}$. The subscript $a$ selects the remaining
$n_v-6$ actuated rows and columns, and the joint torque is recovered from
them as
$\tau(\dot{v},f)=M_a\dot{v}+\eta_a-J_{c,a}^\top f\in\mathbb{R}^{n_v-6}$,
with $|\cdot|$ elementwise. The weight $W\succ0$ scales the acceleration task,
$\bar{\tau}$ is the per-joint torque limit and $\varepsilon>0$ the force
regularization weight. The benchmark uses
$W=\mathrm{diag}(10\cdot\mathbf{1}_6,\mathbf{1}_{n_v-6})$, weighting the
floating-base rows above the joints; the remaining values are given in
Sect.~\ref{sec:protocol}. This QP is small, a few tens of variables, and dense. It must be re-solved at
$100$--$1000$\,Hz.

Equation~\eqref{eq:wbc} is the \emph{weighted} whole-body QP, in which
competing objectives are traded off through $W$. A second family imposes
strict task priorities instead, solving a cascade of QPs in which each level
is restricted to the optimal set of the level above, through null-space
projection or through slack variables on the lower-priority
rows~\cite{kanoun2011,escande2014,herzog2016}; hierarchical controllers of
this kind are standard on quadrupeds~\cite{bellicoso2017,sherikov2014}. The
consequence for solver selection is arithmetic rather than algorithmic: a
$k$-level hierarchy presents the solver with $k$ small dense QPs per control
tick rather than one, so the per-call overhead of Table~\ref{tab:overhead}
and the warm-start behavior of Table~\ref{tab:wsctl} apply $k$ times per
tick. We benchmark the single-QP form; the per-solve numbers transfer to each
level, the per-tick budget does not.

Both templates also assume the contact set is decided upstream. When contact
timing is itself optimized, the problem acquires complementarity conditions
between contact force and gap~\cite{stewart1996} and ceases to be a QP:
contact-implicit trajectory optimization~\cite{posa2014} and contact-implicit
MPC~\cite{lecleach2024} solve mathematical programs with complementarity
constraints, for which none of the solvers in
Table~\ref{tab:qp-solvers} applies. This review is confined to the
fixed-contact-set case.

Equations~\eqref{eq:mpc} and~\eqref{eq:wbc} are the two instances benchmarked
in Sect.~\ref{sec:own-bench}.

\paragraph{Solver families in robotics.}
Four major algorithmic families are commonly used to solve
Eq.~\eqref{eq:qp} in robotics:
\textbf{active-set methods}, \textbf{interior-point methods},
\textbf{operator-splitting methods} (e.g., ADMM), and \textbf{augmented-Lagrangian} and \textbf{proximal methods}~\cite{ProxQP}.
In addition, \textbf{code generation} tools can produce specialized C solvers for fixed-structure problem families, which is often attractive for embedded deployment.

\textbf{Active-set methods}~\cite{qpOASES} iteratively identify the active inequality constraints ($C_{\mathcal{A}} x = d_{\mathcal{A}}$) and solve equality-constrained subproblems:
\begin{equation}
\begin{aligned}
\min_{x \in \mathbb{R}^n} \quad & \frac{1}{2} x^\top H x + q^\top x \\
\text{s.t.} \quad & A x = b, \\
& C_{\mathcal{A}} x = d_{\mathcal{A}},
\end{aligned}
\end{equation}
where $\mathcal{A}\subseteq\{1,\dots,n_i\}$ is the active set. These methods are effective for small-to-medium dense QPs when the active set changes slowly, enabling strong warm-starting. Representative solvers include \texttt{qpOASES}~\cite{qpOASES}, \texttt{Eiquadprog}~\cite{Eiquadprog}, \texttt{qpmad}~\cite{qpmad}, and embedded MPC-oriented dual active-set solvers such as \texttt{DAQP}~\cite{daqp}.
For a safety case the relevant property is a bound on the worst solve, not the
median: active-set methods admit exact complexity certification, in that the
iteration count over a given parameter set can be computed
offline~\cite{cimini2017,arnstrom2022}, which is what \texttt{DAQP} is built
around. At the limit, explicit MPC~\cite{bemporad2002} removes the online
solve altogether in exchange for an offline partition that scales poorly in
the dimensions of Eq.~\eqref{eq:mpc}. We report 99th-percentile times rather
than certified bounds, and the distinction matters when the deadline is hard.

\textbf{Interior-point methods (IPMs)}~\cite{HPIPM} approach the feasible region from the interior by introducing a barrier term (assuming inequalities are expressed as $C x \le d$):
\begin{equation}
\begin{aligned}
\min_{x \in \mathbb{R}^n} \quad &
\frac{1}{2} x^\top H x + q^\top x - \sigma \sum_{i=1}^{n_i} \log\!\left(d_i - (C x)_i\right) \\
\text{s.t.} \quad & A x = b,
\end{aligned}
\end{equation}
where $\sigma>0$ is the barrier parameter and strict feasibility $C x < d$ is required. Structured IPMs such as \texttt{HPIPM}~\cite{HPIPM} exploit OCP sparsity in MPC; sparse robotics-oriented IPMs include \texttt{qpSWIFT}~\cite{qpSWIFT}. Proximal IPM variants such as \texttt{PIQP}~\cite{piqp} target ill-conditioned sparse QPs while retaining embedded-friendly implementation traits. Outside robotics, \texttt{NASOQ}~\cite{nasoq} was designed on the premise that general sparse QP solvers do not reach, at default settings, the accuracy contact-rich physics requires, which is the effect we measure in Table~\ref{tab:accuracy} on a different problem population.

\textbf{ADMM-based solvers}~\cite{OSQP} introduce slack variables for inequalities. For $C x \le d$, define $s\in\mathbb{R}^{n_i}$ such that $C x + s = d,\; s \ge 0$. Then:
\begin{equation}
\begin{aligned}
\min_{x \in \mathbb{R}^n,\, s \in \mathbb{R}^{n_i}} \quad &
\frac{1}{2} x^\top H x + q^\top x + I_{\mathbb{R}^{n_i}_{+}}(s) \\
\text{s.t.} \quad &
A x = b,\;\;
C x + s = d,
\end{aligned}
\end{equation}
where $I_{\mathbb{R}^{n_i}_{+}}(s)$ is the indicator function of the nonnegative orthant. Operator-splitting methods can provide predictable per-iteration cost and effective warm-starting; \texttt{OSQP}~\cite{OSQP} is a widely used representative.

\textbf{Augmented Lagrangian/proximal methods}~\cite{ProxQP} minimize a penalized Lagrangian. A standard (Powell--Hestenes--Rockafellar) form for $C x \le d$ is, with $\max$ taken elementwise:
\begin{equation}
\begin{split}
\mathcal{L}_{\rho_e,\rho_i}(x, \lambda, z) =\;&
\frac{1}{2} x^\top H x + q^\top x
+ \lambda^\top (A x - b)
+ \frac{\rho_e}{2} \| A x - b \|_2^2 \\
&+ \frac{\rho_i}{2} \left\|\max\!\left(0,\, C x - d + \tfrac{1}{\rho_i}z\right)\right\|_2^2
- \frac{1}{2\rho_i}\|z\|_2^2,
\end{split}
\end{equation}
with $\lambda\in\mathbb{R}^{n_e}$, $z\in\mathbb{R}^{n_i}$ and $\rho_e,\rho_i>0$. A typical dual update, at inner iteration $\ell$, is:
\begin{align}
\lambda^{\ell+1} &= \lambda^{\ell} + \rho_e (A x^{\ell+1} - b), \\
z^{\ell+1} &= \max\!\bigl(0,\, z^{\ell} + \rho_i (C x^{\ell+1} - d)\bigr).
\end{align}
\texttt{ProxQP}~\cite{ProxQP} is a representative solver used in robotics that often exhibits strong robustness in contact-rich settings.

\textbf{Code generation and conic solvers.}
When the QP structure is fixed (dimensions and sparsity pattern constant) and only parameters change online, code-generation tools such as \texttt{CVXGEN}~\cite{cvxgen}, \texttt{CVXPYgen}~\cite{cvxpygen}, and \texttt{FORCESPRO}~\cite{forcespro} can produce specialized C solvers with tight footprints and predictable timing.
A related embedded route keeps a fixed-structure solver instead of generating one: \texttt{TinyMPC}~\cite{tinympc} specializes ADMM to linear MPC by precomputing and caching the Riccati terms offline, removing the online factorization so that the remaining solver data fits in a microcontroller's memory.
When the formulation involves conic constraints (beyond standard QP inequalities) or certificates are desired, conic IPMs such as \texttt{Clarabel}~\cite{clarabel} provide a general convex optimization route that includes QPs as a special case, albeit with potential overhead compared to QP-specialized solvers.

\texttt{ECOS}, \texttt{CVXOPT} and \texttt{HiGHS} also enter the
measurements of Sect.~\ref{sec:own-bench}, because they are reachable through
the same interface, although none is designed for this workload:
respectively an embedded conic IPM~\cite{ecos}, a general
convex-optimization package~\cite{cvxopt}, and an LP and mixed-integer
code~\cite{highs} whose convex-QP capability is a later active-set addition,
the most likely reason its per-call floor in Table~\ref{tab:overhead} is the
largest in the set.

Table~\ref{tab:qp-solvers} collects these families and their representative
solvers in one place, with the scenarios each is suited to and the
limitation that most often bites in practice. It is the qualitative
counterpart to the measurements of Sect.~\ref{sec:own-bench}; where the two
disagree, the measurements win, and we note those cases as they arise.

\begin{landscape}
\begin{center}
\scriptsize

\setlength{\tabcolsep}{3pt}
\begin{longtable}{@{}p{2.7cm}p{2.9cm}c c c c p{4.7cm}p{5.0cm}@{}}
\caption{Unified overview of QP solver \emph{families} and representative \emph{solvers} for real-time robotics.
AS = active set; AL = Augmented Lagrangian; IP = Interior-Point; OS = Operator Splitting; 
WS = warm-start; Acc = high-accuracy capability (when configured accordingly);
Early = early termination / inexact solve support; RT = practical real-time suitability (scenario-dependent).}
\label{tab:qp-solvers}\\
\toprule
\textbf{Family / Solver} & \textbf{Method family} & \textbf{WS} & \textbf{Acc} & \textbf{Early} & \textbf{RT} & \textbf{Best scenarios} & \textbf{Key notes (advantages / limitations)} \\
\midrule
\endfirsthead
\toprule
\textbf{Family / Solver} & \textbf{Method family} & \textbf{WS} & \textbf{Acc} & \textbf{Early} & \textbf{RT} & \textbf{Best scenarios} & \textbf{Key notes (advantages / limitations)} \\
\midrule
\endhead

\multicolumn{8}{@{}l}{\textbf{(A) Algorithm-family summary}}\\
\midrule
Direct (eq.-only) & KKT / linear solve & N/A & \cmark & \xmark & \cmark &
Equality-only QPs; fixed structure & Very fast where applicable, but robotics almost always needs inequalities (contacts, friction, torque limits).\\

AS & AS (primal/dual) & \cmark & \cmark & \xmark & \cmark &
Small--medium dense QPs (WBC, condensed MPC) & Strong warm start if the active set moves slowly; timing jitter at contact switches.\\

IP (generic) & Primal-dual IPM & \xmark & \cmark & \cmark & \xmark &
Offline/high-accuracy validation; general sparse QPs & Robust and accurate; heavy for hard real-time unless structure is exploited.\\

IP (structured) & Structure-exploiting IPM & \cmark & \cmark & \cmark & \cmark &
Sparse OCP/MPC QPs (block-banded/tree) & Real-time once OCP sparsity is exploited; little benefit if the structure does not match.\\

AL / proximal & AL / proximal Newton & \cmark & \cmark & \cmark & \cmark &
Contact-rich WBC/MPC; occasional infeasibility & Robust in practice and supports inexact solves; needs penalty and scaling tuning.\\

OS & ADMM / splitting & \cmark & \xmark & \cmark & \cmark &
Large sparse QPs; repeated solves; moderate accuracy & Predictable iteration cost; slow to high accuracy; tuning matters.\\

GPU-resident & ADMM / IPM / iLQR on device & \cmark & \cmark & \cmark & \cmark &
Long-horizon MPC; batched solves; sampling-based control & Large gains once one solve fills the device, or across a batch. A per-call round trip exceeds a whole WBC solve, so single small QPs lose (Sect.~\ref{sec:gpu}).\\

\midrule
\multicolumn{8}{@{}l}{\textbf{(B) Representative solvers (selection-oriented)}}\\
\midrule
\texttt{HPIPM}~\cite{HPIPM} &
IP, structure-exploiting &
\cmark & \cmark & \cmark & \cmark &
MPC / OCP QPs, tree-structured dynamics &
Very fast when the OCP structure matches; embedded focus. Weaker on arbitrary sparse QPs.\\

\texttt{OSQP}~\cite{OSQP} &
ADMM / OS &
\cmark & \xmark & \cmark & \cmark &
Large sparse QPs; repeated solves with warm-start &
Sparse-first, small footprint, factorization caching. Loose default tolerance (Table~\ref{tab:accuracy}).\\

\texttt{ProxQP}~\cite{ProxQP} &
AL / proximal &
\cmark & \cmark & \cmark & \cmark &
WBC, inverse dynamics, small--medium QPs &
Strong robotics results and a good accuracy trade-off. Check \texttt{eps\_primal\_inf} on condensed problems (Table~\ref{tab:proxqp}).\\

\texttt{qpOASES}~\cite{qpOASES} &
Parametric AS &
\cmark & \cmark & \xmark & \cmark &
Small/medium dense MPC/WBC with stable active sets &
Excellent warm start, fast when the active set is stable; worst-case iteration count varies.\\

\texttt{qpmad}~\cite{qpmad} &
Dual AS (Goldfarb--Idnani) &
\cmark & \cmark & \cmark & \cmark &
Dense WBC; condensed MPC dense QPs &
Lightweight, low-overhead C++; strong on repeated dense solves; spikes on abrupt active-set changes.\\

\texttt{DAQP}~\cite{daqp} &
Dual AS&
\cmark & \cmark & \cmark & \cmark &
Fully condensed MPC dense QPs &
Built for embedded MPC; worst case boundable offline. Fastest here on dense WBC; poor on sparse (Table~\ref{tab:timing-sparse}).\\

\texttt{PIQP}~\cite{piqp} &
Proximal IP &
\cmark & \cmark & \cmark & \cmark &
Ill-conditioned sparse QPs; embedded constraints &
Handles ill-conditioning without strong constraint qualifications. Fastest here on sparse OCP QPs; slower once condensed.\\

\texttt{qpSWIFT}~\cite{qpSWIFT} &
Sparse primal-dual IPM &
\cmark & \cmark & \cmark & \cmark &
Embedded sparse robotics QPs &
Lightweight sparse IPM with sparse LDL. Less common in mainstream stacks; integration cost may dominate.\\

\texttt{FORCESPRO}~\cite{forcespro} &
Codegen + IPM / fast QP modes &
\cmark & \cmark & \cmark & \cmark &
Hard real-time embedded MPC with tooling &
Mature code generation and deployment tooling. Commercial license; less transparent than open source.\\

\texttt{CVXGEN}~\cite{cvxgen} / \texttt{CVXPYgen}~\cite{cvxpygen} &
Code generation (fixed structure) &
N/A & \cmark & \xmark & \cmark &
Very small fixed-structure QPs &
Flat C code for a fixed problem family; excellent for tiny fixed QPs, inflexible if the formulation changes.\\

\texttt{Clarabel}~\cite{clarabel} &
IP (conic; QP-capable) &
\cmark & \cmark & \cmark & \cmark &
When conic constraints/certificates are needed &
General conic solver with quadratic objectives. Generality costs overhead in tight loops.\\

\bottomrule
\end{longtable}

\end{center}
\end{landscape}
\setlength{\tabcolsep}{6pt}
\section{Evidence from Public Benchmarks}
\label{sec:public}
{\small
\begin{longtable}{@{}l l c c c c c@{}}
\caption{Interface-level feature comparison of representative real-time QP solvers.
\textbf{WS} = accepts a primal--dual warm start through its public API;
\textbf{HS} = hot start, i.e. the factorization and/or active set can be reused when only
the problem \emph{data} change but the sparsity pattern does not;
\textbf{ET} = a user-settable tolerance allows early termination at reduced accuracy;
\textbf{DG} = the termination criterion contains an explicit duality-gap test;
\textbf{IC} = the solver \emph{detected} infeasibility, i.e. refused to return
a solution, on the test of Sect.~\ref{sec:results}; we did not test for a
Farkas certificate, so the column says less than its name suggests.
Entries marked $^{\dagger}$ are solvers measured in
Sect.~\ref{sec:own-bench}, and their \textbf{IC} entry is what we observed on
a deliberately infeasible WBC problem; for the remaining solvers the
\textbf{IC} column is left as ``---'' because we did not test them, and we do
not restate a claim we could not check. The other columns follow the solver
documentation.} \label{tab:qpchars} \\
\toprule
\textbf{Solver (Method)} & \textbf{Matrix} & \textbf{WS} & \textbf{HS} & \textbf{ET} & \textbf{DG} & \textbf{IC} \\
\midrule
\endfirsthead
\toprule
\textbf{Solver (Method)} & \textbf{Matrix} & \textbf{WS} & \textbf{HS} & \textbf{ET} & \textbf{DG} & \textbf{IC} \\
\midrule
\endhead
\texttt{qpOASES} (AS)\cite{qpOASES}         & Dense / Sparse    & \cmark & \cmark & \xmark & \cmark & --- \\
\texttt{quadprog} (AS)$^{\dagger}$\cite{ref_goldfarb1983}        & Dense             & \xmark & \xmark & \xmark & \cmark & \cmark \\
\texttt{Eiquadprog} (AS)\cite{Eiquadprog}      & Dense             & \cmark & \xmark & \xmark & \cmark & --- \\
\texttt{qpmad} (AS)\cite{qpmad}           & Both              & \cmark & \cmark & \cmark & \cmark & --- \\
\texttt{DAQP} (AS)$^{\dagger}$\cite{daqp}           & Both              & \cmark & \cmark & \cmark & \cmark & \cmark \\
\texttt{Gurobi} (IP)\cite{Gurobi}          & Sparse            & \xmark & \xmark & \cmark & \cmark & --- \\
\texttt{MOSEK} (IP)\cite{MOSEK}           & Sparse            & \xmark & \xmark & \cmark & \cmark & --- \\
\texttt{HPIPM} (IP)\cite{HPIPM}           & Sparse (OCP form) & \cmark & \cmark & \cmark & \cmark & --- \\
\texttt{PIQP} (IP)$^{\dagger}$\cite{piqp}            & Both              & \cmark & \cmark & \cmark & \cmark & \cmark \\
\texttt{Clarabel} (IP)$^{\dagger}$\cite{clarabel}    & Sparse            & \cmark & \cmark & \cmark & \cmark & \cmark \\
\texttt{OSQP} (OS)$^{\dagger}$\cite{OSQP}            & Sparse            & \cmark & \cmark & \cmark & \xmark & \cmark \\
\texttt{SCS} (OS)$^{\dagger}$\cite{SCS}             & Sparse            & \cmark & \cmark & \cmark & \xmark & \cmark \\
\texttt{QPALM} (AL)$^{\dagger}$\cite{qpalm}          & Both              & \cmark & \cmark & \cmark & \cmark & \cmark \\
\texttt{ProxQP} (AL)$^{\dagger}$\cite{ProxQP}          & Both              & \cmark & \cmark & \cmark & \cmark & \cmark \\
\bottomrule
\end{longtable}
}
The public efforts that bear directly on legged robotics are restated here
strictly as those sources report them.

\paragraph{The \texttt{qpbenchmark} suite.}
\texttt{qpbenchmark}~\cite{qpbenchmark} ships three test sets: a
\emph{free-for-all} collection, the \emph{Maros--Meszaros} problems, which are
designed to be numerically difficult, and a \emph{model predictive control}
set. The MPC set is the closest public analogue to what a legged robot solves
and contains three families: quadruped centroidal MPC, linear-inverted-pendulum
humanoid walking, and wheeled-inverted-pendulum balancing.
\texttt{qpbenchmark} does report primal residual, dual residual and duality
gap alongside runtime, at a default and a high-accuracy settings preset, and
aggregates runtime and residuals by shifted geometric mean in the manner of
performance profiles~\cite{dolan2002}. What it does not provide is a
\emph{generator}: its instances are fixed artifacts, none drawn
from the two templates of Eqs.~\eqref{eq:mpc} and~\eqref{eq:wbc} in both
their sparse and condensed forms, and its aggregation is over a heterogeneous
population, not over one control loop.

\paragraph{Quadrupedal walking: Stark et al.}
The study closest to the present paper~\cite{stark2025benchmarking} benchmarks
six solvers, \texttt{HPIPM} (in two modes, \texttt{balanced} and
\texttt{speed\_abs}, treated as two solvers), \texttt{OSQP}, \texttt{qpOASES},
\texttt{DAQP}, \texttt{Eiquadprog} and \texttt{PROXQP}, on two canonical
control problems: MPC for trajectory optimization, implemented through
\texttt{acados}~\cite{acados}, and whole-body inverse dynamics, implemented in
the ARC-OPT stack. Its organizing axis is \emph{partial condensing}: the
authors sweep every condensing level from fully sparse to fully dense, on three
platforms: a desktop i9 CPU, a LattePanda Alpha, and an NVIDIA Jetson Orin
NX. All solvers are run at default hyper-parameters, deliberately, ``to ensure
better comparability''.

For MPC the authors evaluate \texttt{HPIPM} (both modes), \texttt{OSQP},
\texttt{qpOASES} and \texttt{DAQP}, and report that for \emph{all}
architectures, condensing levels and horizons the best-performing solver is
\texttt{HPIPM} in \texttt{speed\_abs} mode. They attach an important caveat to
that result: \texttt{speed\_abs} ``does not provide accurate results for some
applications''. The active-set solvers \texttt{qpOASES} and \texttt{DAQP} are
competitive only at the short horizon $N=10$, where they nearly match
\texttt{OSQP}, and fall well behind at $N=20$; the authors suggest active-set
methods would be advantageous below $N=10$, and note that their MPC problem
``seems to be at the limit of the problem size where ASM is outperformed''.

For WBC the authors evaluate \texttt{Eiquadprog}, \texttt{PROXQP},
\texttt{qpOASES} and \texttt{HPIPM}. \texttt{Eiquadprog} is fastest in every
comparison, but the headline is the spread, not the ranking: all four
solvers stay below $0.5$\,ms average cycle time on every platform. The authors
conclude that ``in WBC, any of the regarded open-source solvers performs well;
the engineering effort should rather be put into the formulation of the WBC
problem itself''. We stress this because it cuts against the instinct to shop
for a WBC solver: on their evidence the formulation, not the solver, is where
the engineering effort belongs.

The study also introduces the Solve Frequency per Watt metric, adapted from the
Green500 FLOPS-per-Watt measure. The Jetson Orin NX achieves the highest SFPW
on every test, more than twice the LattePanda and about three times the
desktop, with \texttt{HPIPM} \texttt{speed\_abs} the most efficient solver
for MPC on all systems and \texttt{Eiquadprog} for WBC. Two things the study
does \emph{not} report are worth naming, because they bound what can be
concluded from it: it compares no solution accuracy between solvers, and it
performs no solver-failure or robustness analysis.

\paragraph{Dense solvers: \texttt{qpmad\_benchmark}.}
A complementary, narrower evaluation~\cite{qpmad_benchmark} compares
\texttt{qpmad}~\cite{qpmad} against \texttt{qpOASES} and
\texttt{eiquadprog-fast} on dense, positive-definite problems in two sets: the
standalone problems distributed with \texttt{qpOASES}, solved cold, and the MPC
sequences distributed with \texttt{qpOASES}, which ``all have constant Hessian
and varying constraints'' and are hot-started from the second problem onward.
The repository publishes per-problem timing plots instead of any aggregate
speed-up figure, so we quote none; it does record that
\texttt{eiquadprog} ``does not produce the exact answer (up to 1e-9 precision)
on some of the problems''. It also notes, and we return to this in
Sect.~\ref{sec:results}, that \texttt{qpOASES} is used there ``with MPC
option preset'', which ``gives \textasciitilde50\% performance boost compared
to default preset''. Beyond the benchmark, \texttt{qpmad} has been adopted as
the QP back end of a field-deployed robotic perception
pipeline~\cite{qpmad_app}, which is evidence of robustness in integration even
though that study reports no control-loop timings.

\begin{figure}[tbp]
    \centering
    \includegraphics[width=0.85\textwidth]{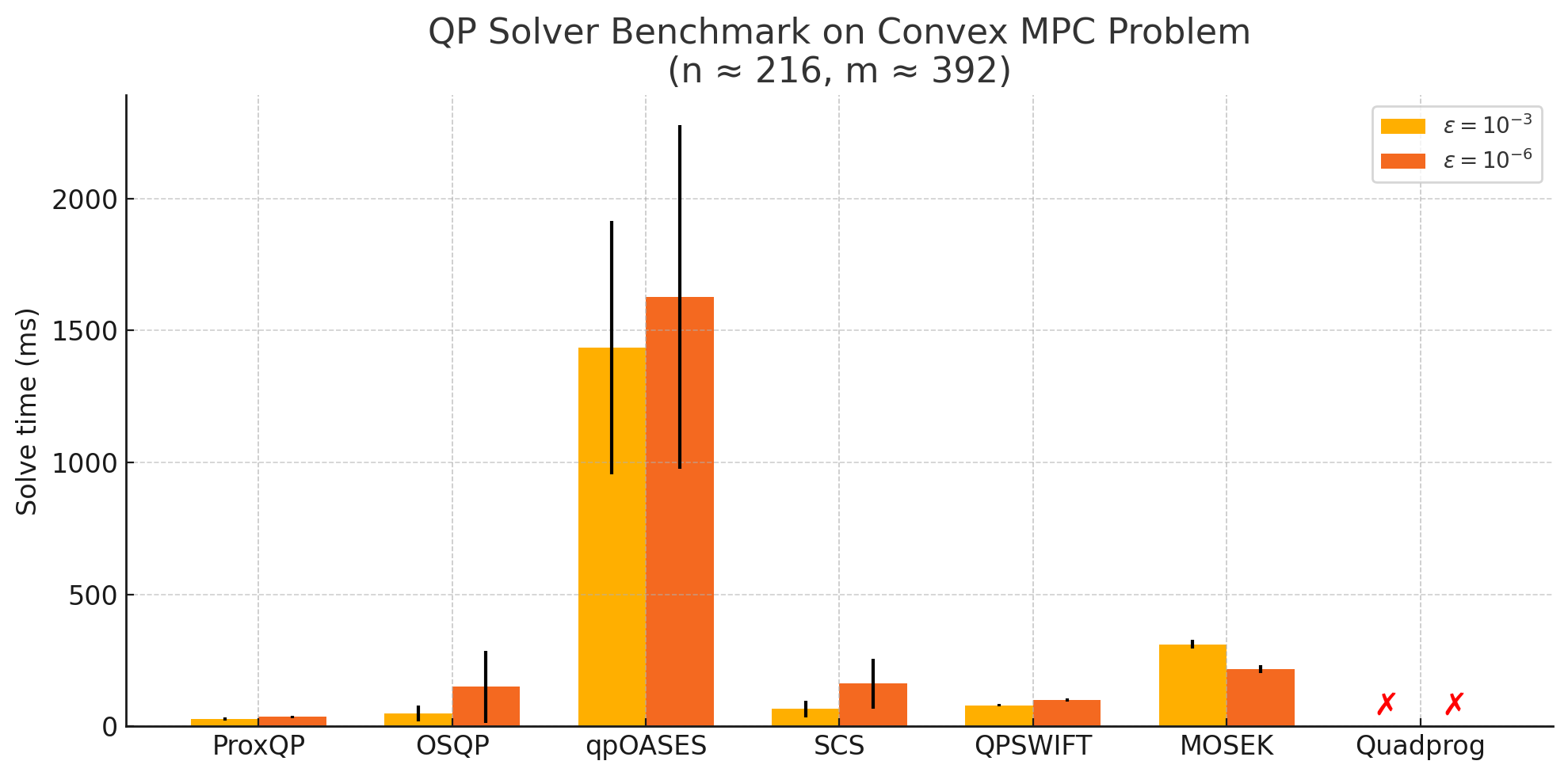}
    \caption{Mean solve time (error bars: one standard deviation) of seven QP
    solvers on a convex MPC instance with $n\approx216$ and
    $n_e+n_i\approx392$, at a loose and a tight tolerance. A ``\xmark'' marks
    a solver that returned no solution at that tolerance. This is a
    single-instance illustration used to motivate
    Sect.~\ref{sec:own-bench}; unlike the measurements reported there, it is
    not accompanied by a released generator and harness, so it should be read
    qualitatively.}
    \label{fig:qp_solver}
\end{figure}

\paragraph{A motivating measurement.}
Fig.~\ref{fig:qp_solver} shows a measurement of our own on a single
convex MPC instance with $n\approx216$ decision variables and
$n_e+n_i\approx392$ constraints: for each of seven solvers, the mean solve
time over a 100-step simulation, with one standard deviation as an error bar,
at a loose ($\epsilon=10^{-3}$) and a tight ($\epsilon=10^{-6}$) tolerance.
We use it here to motivate, not to conclude, because it exhibits in
miniature the three effects that Sect.~\ref{sec:own-bench} then measures
systematically. The spread across solvers on one instance covers almost two
orders of magnitude, so an aggregate ranking over a heterogeneous suite can
hide the behavior that matters for one control loop. Tightening the tolerance
costs the operator-splitting solvers far more than the interior-point ones.
And one solver returns nothing at either tolerance, a reminder that a
missing bar is a result, not a gap in the data.

\paragraph{What the public evidence leaves open.}
Taken together these studies establish a clear structural story: exploit OCP
sparsity for long-horizon MPC, and expect dense active-set methods to be
adequate for WBC. What they do not settle, with the partial exception of
\texttt{qpbenchmark}, which does report residuals on its own heterogeneous
population, is how accurate the returned solutions are on legged-robot
instances, how much of a solver's advertised warm-start benefit is reachable
in practice, and how the solvers behave when their default settings meet a
problem scaling their authors did not anticipate. Those are the
questions Sect.~\ref{sec:own-bench} measures.

\input{own_bench}

\input{gpu}

\input{guide}

\section{Discussion and Conclusion}

We reviewed the QP solver families used in real-time legged-robot control,
consolidated the public benchmarking evidence for them, and added a
reproducible benchmark on the two QP templates that a legged-robot stack
actually solves.

The review part of the picture is largely confirmed by the measurements.
Structure-exploiting and sparse interior-point solvers dominate long-horizon
MPC in OCP form; dense active-set methods dominate the small dense QPs of
whole-body control, where \texttt{qpmad} and \texttt{eiquadprog} solve the
inverse-dynamics problem in under $0.03$\,ms natively, leaving more than an
order of magnitude of headroom at $1$\,kHz; operator-splitting and proximal
augmented-Lagrangian solvers occupy the middle ground where moderate accuracy
and predictable iteration cost are the priority; code generation remains the
answer when the formulation is frozen and the target is a microcontroller; and
conic solvers broaden the modeling scope at a cost in tight loops.

The measurements add three things that a solve-time ranking does not show.
Each of them can break a deployment. The first is the
\emph{formulation}: a structural mismatch between the QP and the solver cost
us up to $31\times$ on one and the same optimization problem, far more than
the spread between sensible solvers within the right structure, and whether
condensing helps or hurts flips sign between $N=10$ and $N=20$. The second is
the \emph{default settings}. One solver's
default options are $228\times$ slower than its own shipped MPC preset on the
same problem; another's default infeasibility threshold turns a feasible
condensed MPC problem into an immediate, confident and wrong infeasibility
certificate; and the operator-splitting solvers stop many orders of magnitude
earlier than the interior-point ones. None of these is a defect in the
algorithms. Each is a default chosen for a problem population that a legged
robot's QPs do not belong to. The third is the \emph{interface}: separating setup reuse
from warm starting shows that keeping the solver object alive is worth far
more than the warm start itself for a dual active-set method, while the warm
start proper is worth most to the proximal and operator-splitting solvers.
Neither is reachable through a generic wrapper, and a benchmark conducted
through one will rank solvers in an order that does not survive
deployment.

The main gaps are stated in Sect.~\ref{sec:limits}: no energy measurement of
our own; no coverage of \texttt{HPIPM} or \texttt{qpSWIFT}, which would need
a stage-wise OCP harness of their own; and a WBC comparison that reaches a
real robot only through one model in two postures. Closing the first two on a
robot's own onboard computer, with instrumented power measurement and a
structured-OCP interface, is the natural next step.

Finally, Sect.~\ref{sec:gpu} places the GPU question for legged control. The
GPU-MPC literature and our own floor measurement agree once they are read as
statements about different regimes: a GPU pays for long horizons, for batched
solves and for sampling-based control, and cannot pay for a single
thirty-variable QP whose entire CPU solve is shorter than one round trip to
the device. That boundary falls between the MPC and WBC layers and not
around them, which is a more useful thing for a system designer to know than
any single speed-up figure.

\subsubsection*{Code availability.}
The problem generator, the Python and C++ measurement harnesses, the GPU
probe, the seed study of Sect.~\ref{sec:limits}, the raw results and the
scripts that emit every table and figure in
Sects.~\ref{sec:own-bench} and~\ref{sec:gpu} are released with this paper as
ancillary files under the MIT license, with a pinned
\texttt{requirements.txt}; \texttt{run\_all.sh} regenerates them end to end
and exports the thread and core-affinity settings used for the reported
timings. Two caveats on reproducibility remain. The C++ dependencies are
cloned from upstream and, in the run reported here, were
not pinned to a commit (the script now records and accepts refs), and three
Python solver versions were not captured by that run. Those gaps are noted in
\texttt{requirements.txt}.

\end{document}

%% file: own_bench.tex
\section{A Reproducible Benchmark on Legged-Robot QPs}
\label{sec:own-bench}

This section answers those questions with a small, fully reproducible
benchmark built directly on Eqs.~\eqref{eq:mpc} and~\eqref{eq:wbc}.

\subsection{Protocol}
\label{sec:protocol}

The generator, both measurement harnesses and the table/figure script are
included with this submission as ancillary files (\texttt{anc/}), together
with the raw results of the run reported here; \texttt{run\_all.sh}
regenerates everything in this section end to end. It includes
\texttt{proxqp\_version\_check.py}, a one-command re-check of the finding of
Sect.~\ref{sec:results} against whatever \texttt{proxsuite} a reader has
installed, so that the claim can be confirmed or refuted on another version
and platform without re-running the benchmark.

\paragraph{Instances.}
Every instance is produced by the released generator from a fixed random
seed, so no data files are needed to reproduce them. The two families are
built differently, and the difference bounds what may be concluded from each.

The MPC instances are \emph{analytic}. The matrices $A_k$ and $B_k$ of
Eq.~\eqref{eq:mpc} are the single-rigid-body dynamics of a quadruped with the
following parameters: mass $12$\,kg, body inertia
$\mathrm{diag}(0.07,0.26,0.24)$\,kg\,m$^2$, four contacts at the nominal
stance geometry, friction coefficient $\mu=0.6$, $f_z\in[5,250]$\,N and
$\mathrm{d}t=30$\,ms. A single linearization is used, so $A_k$ and $B_k$ are constant over the
horizon; only the initial state and the reference vary with the seed. Instances are built at horizons $N\in\{10,20,30,40\}$, in both the
sparse OCP form and the condensed dense form. A ridge of $10^{-6}$ is added
to the state Hessian so the problem is strictly convex, which several dual
active-set solvers require.

The WBC instance is \emph{structural} rather than analytic, and we are
explicit about it. Its dimensions, constraint structure and scaling are those
of a quadruped in stance: $n_v=18$ generalized accelerations and
$3n_c=12$ contact forces, giving $n=30$; six unactuated floating-base rows
plus twelve contact-acceleration rows, $n_e=18$; a four-sided friction
pyramid and normal-force bounds per foot plus two-sided torque limits
$\bar{\tau}=33.5$\,N\,m, $n_i=48$; forces in newtons against a $250$\,N
bound. But the mass matrix $M$, the nonlinear terms $\eta$, the contact Jacobian
$J_c$ and the desired acceleration $\dot{v}^{\mathrm{des}}$ are drawn from the
seeded random number generator, not evaluated from a URDF. $M$ is
constructed as $LL^\top+n_v I$, so it is symmetric positive definite and well
conditioned. The force regularization is
$\varepsilon=10^{-4}$. What this instance reproduces faithfully is the
\emph{shape} of a WBC problem --- its size, its constraint types, its units;
what it does not reproduce is a stance. Because $\eta$ is random rather than
containing $mg$, the floating-base rows never require the feet to carry the
body. At the optimum the four normal forces total $21.7$\,N against the
$117.7$\,N weight of the quadruped modeled on the MPC side, three of the four
sit on their lower bound, and twelve of the $48$ inequalities are active. A robot
in stance presents a different active set, and since active-set cost is driven
by the distance from the cold start to that set, the WBC \emph{ordering} below
should be read as a result about a dense QP of this size, structure and
scaling, not as a ranking for a particular robot. Sect.~\ref{sec:limits} says what follows from that.

\paragraph{Solvers.}
Every solver reachable through \texttt{qpsolvers}~4.13 on the test machine is
included, with its default settings: Clarabel, CVXOPT, DAQP, ECOS, HiGHS,
OSQP, PIQP, ProxQP, QPALM, \texttt{quadprog} and SCS. Defaults are the point:
a practitioner integrating a solver into a control stack almost always starts
there, so the behavior of the defaults is what the guidance in
Sect.~\ref{sec:guide} has to be built on.

\texttt{qpOASES}, \texttt{qpmad} and \texttt{eiquadprog} are not reachable
through that interface on the test machine without building them from source,
so we measure them in a second, native C++ harness that reads the identical
problem data.
\texttt{Gurobi} and \texttt{MOSEK}, which \texttt{qpsolvers} also supports,
are omitted for want of a license, not of an interface; their rows in
Table~\ref{tab:qpchars} therefore rest on documentation, not measurement. On the cold and
preset rows the agreement is exact, every C++ solver returning the optimal
values of Table~\ref{tab:proxqp} to all printed digits, which is the
correctness check on the harness itself. The four hot-start rows re-solve
after perturbing the linear term and so return the optimum of a different
problem, differing in the fourth to sixth significant figure.

\paragraph{Measurement.}
The host is an Intel Core i7-13650HX (6 performance cores at $4.9$\,GHz and 8
efficiency cores, 20 threads) running Linux 7.0 with Python 3.11 and
\texttt{qpsolvers}~4.13; solver versions are recorded in the released
\texttt{results.json}. Each measurement is the median of 100 timed solves (400
for the WBC problem) after three warm-up solves, pinned to the performance
cores with a single BLAS thread. We report the median and the 99th percentile,
because a control loop is sized by its tail, not its average. Timings are
taken around the \texttt{qpsolvers} call and therefore include the per-call
Python binding cost; Table~\ref{tab:overhead} quantifies that floor
separately for every solver so it can be subtracted when interpreting the
smallest problems. Dense problems are passed as dense arrays and sparse
problems as sparse matrices, which matters because some wrappers reject
sparse input outright.

\paragraph{Two hosts.}
The measurements come from two machines, and which table comes from which is
recorded in the released JSON rather than left to be inferred. The Linux host
just described produced Tables~\ref{tab:timing-sparse}--\ref{tab:proxqp} and
the GPU floor of Table~\ref{tab:gpu}. An Apple arm64 machine running macOS
and Python~3.11 produced Tables~\ref{tab:matched}, \ref{tab:wsctl}
and~\ref{tab:osqpct} and the robot-model comparison of observation~7. The two
carry identical solver versions --- \texttt{qpsolvers}~4.13.0,
\texttt{OSQP}~1.1.3, \texttt{ProxQP}~0.8.2, \texttt{PIQP}~0.7.0,
\texttt{SCS}~3.3.2, \texttt{Clarabel}~0.11.1, \texttt{QPALM}~1.3.0,
\texttt{ECOS}~2.0.14 and \texttt{CVXOPT}~1.3.3 --- so a ratio measured on one
is comparable with a ratio measured on the other. Absolute milliseconds are
not comparable across the two, and no claim here compares them. The native
C++ results of Table~\ref{tab:native} record no host in their output file,
which is a gap in the released data, not a measurement we can
attribute.

\paragraph{Reported quality.}
Alongside the timings we record the primal residual, dual residual and
duality gap that \texttt{qpsolvers} computes for the returned solution; all
three are in the released \texttt{results.json}. The absolute residual of a problem whose bounds are expressed in newtons is
not comparable to one whose bounds are dimensionless. For the tables we
report two scale-free constraint measures instead: the relative inequality violation
$\|\max(0,Cx-d)\|_\infty/\max(1,\|d\|_\infty)$ and the relative equality
residual $\|Ax-b\|_\infty/\max(1,\|b\|_\infty)$. We report both because they
can differ by orders of magnitude on the same solution, and for a legged
robot it is the equality residual, on the dynamics and contact-consistency
rows, that integrates into drift.

\input{tab_timing}

\begin{figure}[tbp]
\centering
\includegraphics[width=\textwidth]{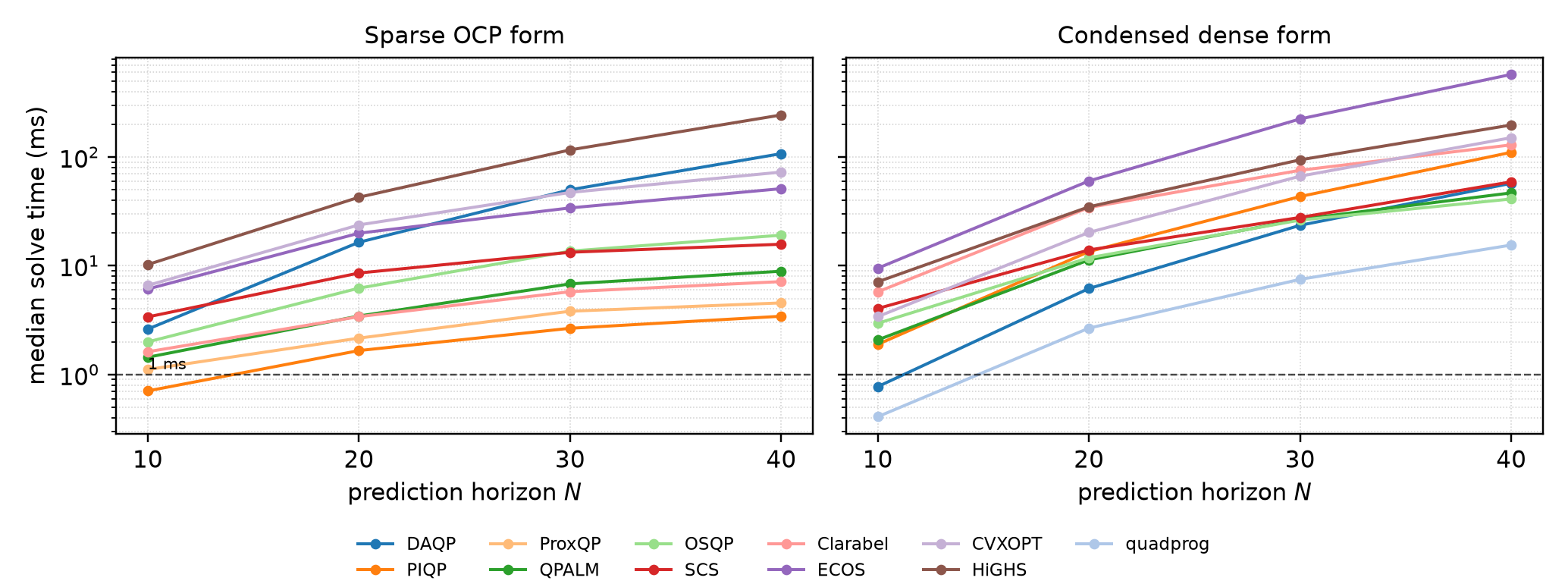}
\caption{Median solve time against prediction horizon for the MPC problem of
Eq.~\eqref{eq:mpc}, in its sparse OCP form (left) and its condensed dense form
(right). Note the logarithmic time axis; the dashed line marks $1$\,ms.
\texttt{ProxQP} is absent from the right panel because it returns no solution
on any condensed instance at its default settings
(Table~\ref{tab:proxqp}).}
\label{fig:horizon}
\end{figure}

\subsection{Results}
\label{sec:results}

\input{results_text}

\input{tab_overhead}
\input{tab_proxqp}

\paragraph{Infeasibility behavior.}
On a deliberately infeasible variant of Eq.~\eqref{eq:wbc}, in which one foot
is required to carry a normal force above its own upper bound, all eleven
solvers behaved correctly, each returning no solution instead of a silently
wrong one. This is the measured basis for the \textbf{IC} column of
Table~\ref{tab:qpchars}; we found no solver in this set that fails to notice a
plainly infeasible WBC problem. The dangerous case is the opposite one, a
\emph{feasible} problem reported as infeasible, which is exactly the
behavior documented in Table~\ref{tab:proxqp}.

\subsection{What this benchmark does not measure}
\label{sec:limits}

Seven limitations bound what the numbers above support.

First, \textbf{no energy measurement}. We had intended to report Solve
Frequency per Watt alongside the timings, following~\cite{stark2025benchmarking}.
On the test machine the Intel RAPL energy counters under
\texttt{/sys/class/powercap} are readable only by \texttt{root}, so no
package-energy figure could be obtained without privileged access, and we
report none instead of estimating one. Every SFPW statement in this paper is
therefore a citation of~\cite{stark2025benchmarking}, not a measurement of
ours.

Second, \textbf{two solvers are still missing}. The native harness closes the
gap for \texttt{qpOASES}, \texttt{qpmad} and \texttt{eiquadprog}, but
\texttt{HPIPM} and \texttt{qpSWIFT} are not measured here. Both are sparse,
and including them would have meant building \texttt{blasfeo} and driving
\texttt{HPIPM}'s stage-wise OCP interface, which is a different harness rather
than another row in Table~\ref{tab:native}. Since
Sect.~\ref{sec:public} reports \texttt{HPIPM} as the fastest solver for sparse
MPC in the published quadrupedal study, and our sparse column is headed by
\texttt{PIQP}, the obvious open question, whether a structure-exploiting
IPM beats a general sparse IPM on these particular instances, is one our
measurements do not answer.

Third, \textbf{the warm-start decomposition rests on one instance family}:
condensed MPC problems at $N=10$ along a closed-loop sequence, measured on
macOS/arm64. The harness's drift floor there is $1.03\times$, so
\texttt{DAQP}'s $1.22\times$ and \texttt{OSQP}'s $1.00\times$ sit at or close to
the level at which an effect can be separated from run-to-run variation. The
ratios are stable across five perturbation sizes, but no step of this sequence
changes the contact set, and we do not claim they transfer to the sparse form
or across a contact switch.

Fourth, \textbf{the WBC instance is synthetic in its dynamics}. As
Sect.~\ref{sec:protocol} states, its mass matrix and contact Jacobian are
random rather than evaluated from a robot model. The timings therefore
characterize how each solver handles a dense QP of that size, constraint
structure and scaling, which is what distinguishes the solvers here, but
they do not capture how a particular robot's conditioning, or a particular
contact configuration, would stress them. A near-singular mass matrix or a
degenerate contact set is exactly the regime in which the differences we
report could change. Observation~7 bounds that cost by rebuilding the same QP
from a real URDF. The ordering moves only in its middle, no solver fails on
either instance, and the robot's own contact Jacobian turns out to be better
conditioned than one might fear; what the synthetic instance does not
reproduce is the stance itself. The MPC instances do not share this
limitation, their dynamics being the analytic SRBD model.

Fifth, \textbf{neither host is a robot's onboard computer}. The Linux host of Tables~\ref{tab:timing-sparse}--\ref{tab:proxqp} is a
laptop-class x86 CPU left under the \texttt{powersave} frequency governor,
which could not be changed without privileged access and which adds
clock-frequency variation on top of the measured dispersion. The macOS host
of Tables~\ref{tab:matched}--\ref{tab:osqpct} is a laptop too, with no
governor control of its own. Absolute milliseconds will differ on embedded ARM
hardware. The ordering between solvers, the accuracy differences, and the
structural effects (sparse versus condensed) are the transferable part, and
the harness is released so that the same numbers can be regenerated on the
target platform.

Sixth, \textbf{the main tables rest on one draw of each instance}. The random
WBC instance and the MPC initial state are generated from a single seed, and
the 99th percentile over $100$ repeats is the second-largest of $100$
observations, a high-variance statistic even though it is the one a control
budget cares about. To bound what that costs, \texttt{seed\_study.py}
repeats the synthetic WBC family, the condensed MPC family and the warm-start
decomposition over five seeds. The decomposition is insensitive to the draw:
across seeds \texttt{DAQP}'s setup reuse stays within $23.4$--$24.0\times$
and its warm start within $1.19$--$1.24\times$, \texttt{ProxQP}'s warm start
within $1.56$--$1.60\times$, and \texttt{OSQP}'s within $1.00$--$1.00\times$.
The orderings are not uniformly so: the first two places are fixed on both
families, but middle ranks move by one or two positions between seeds, and
\texttt{OSQP} on the synthetic WBC instance moves between sixth and eleventh
with a $2.0\times$ spread in its median time. Adjacent solvers separated by
tens of percent in a single table should therefore be read as unresolved; the
orderings this paper draws conclusions from are separated by factors.

Seventh, \textbf{the relative inequality violation is normalized by the
largest bound in the instance}, not row by row. Two thirds of the MPC
inequality rows and a third of the WBC rows are homogeneous friction-pyramid
rows with $d_i=0$, so a violation on one of those is divided by
$\|d\|_\infty=250$ and not by its own scale. The measure is therefore
generous to all solvers alike on those rows; it ranks them consistently but
overstates the absolute quality of every entry in
Table~\ref{tab:accuracy}. The suboptimality of Table~\ref{tab:matched},
which has no such normalization, is the stricter of the two measures.

%% file: tab_timing.tex
\begin{table}[tbp]
\centering
\scriptsize
\caption{Median solve time in milliseconds, 99th percentile in brackets, on the \emph{sparse} OCP form of the MPC problem of Eq.~\eqref{eq:mpc} at horizon $N$, and on the WBC problem of Eq.~\eqref{eq:wbc}. \texttt{n/a}: the wrapper rejects sparse input.}
\label{tab:timing-sparse}
\setlength{\tabcolsep}{3pt}
\begin{tabular}{@{}l l r r r r r@{}}
\toprule
\textbf{Solver} & \textbf{Family} & \textbf{$N=10$} & \textbf{$N=20$} & \textbf{$N=30$} & \textbf{$N=40$} & \textbf{WBC} \\
\midrule
\texttt{DAQP} & dual AS & 2.62\,(3.66) & 16.46\,(18.01) & 50.01\,(52.97) & 107.5\,(111.1) & 0.029\,(0.054) \\
\texttt{quadprog} & dual AS & n/a & n/a & n/a & n/a & 0.159\,(0.508) \\
\texttt{PIQP} & proximal IP & 0.707\,(1.21) & 1.66\,(2.36) & 2.66\,(3.23) & 3.44\,(3.91) & 0.138\,(0.245) \\
\texttt{ProxQP} & AL/prox. & 1.11\,(2.28) & 2.16\,(2.97) & 3.82\,(4.91) & 4.56\,(4.96) & 0.377\,(0.544) \\
\texttt{QPALM} & proximal AL & 1.44\,(2.00) & 3.45\,(5.59) & 6.82\,(9.22) & 8.91\,(9.47) & 0.962\,(1.65) \\
\texttt{OSQP} & ADMM & 2.00\,(2.59) & 6.23\,(7.34) & 13.65\,(15.89) & 19.12\,(20.23) & 0.763\,(1.31) \\
\texttt{SCS} & conic ADMM & 3.39\,(3.77) & 8.58\,(13.13) & 13.32\,(17.80) & 15.78\,(16.28) & 1.13\,(1.72) \\
\texttt{Clarabel} & conic IP & 1.62\,(2.47) & 3.41\,(4.12) & 5.78\,(8.07) & 7.16\,(7.42) & 0.999\,(2.09) \\
\texttt{ECOS} & conic IP & 6.11\,(7.44) & 19.94\,(22.96) & 34.06\,(36.39) & 51.19\,(53.97) & 1.59\,(2.87) \\
\texttt{CVXOPT} & IP & 6.60\,(7.58) & 23.77\,(25.80) & 47.26\,(50.13) & 72.73\,(74.50) & 0.818\,(1.12) \\
\texttt{HiGHS} & IP/simplex & 10.26\,(11.55) & 42.70\,(46.68) & 116.3\,(123.0) & 243.8\,(247.6) & 3.04\,(4.19) \\
\bottomrule
\end{tabular}
\end{table}

\begin{table}[tbp]
\centering
\scriptsize
\caption{Median solve time in milliseconds, 99th percentile in brackets, on the \emph{condensed} dense form of the same MPC problems. \xmark: the solver returned no solution (see Table~\ref{tab:proxqp}).}
\label{tab:timing-cond}
\setlength{\tabcolsep}{3pt}
\begin{tabular}{@{}l l r r r r@{}}
\toprule
\textbf{Solver} & \textbf{Family} & \textbf{$N=10$} & \textbf{$N=20$} & \textbf{$N=30$} & \textbf{$N=40$} \\
\midrule
\texttt{DAQP} & dual AS & 0.774\,(1.20) & 6.18\,(6.76) & 23.57\,(32.77) & 57.33\,(62.21) \\
\texttt{quadprog} & dual AS & 0.409\,(0.640) & 2.67\,(4.67) & 7.51\,(8.19) & 15.52\,(22.91) \\
\texttt{PIQP} & proximal IP & 1.89\,(2.31) & 13.51\,(15.81) & 43.34\,(44.96) & 110.6\,(122.4) \\
\texttt{ProxQP} & AL/prox. & \xmark & \xmark & \xmark & \xmark \\
\texttt{QPALM} & proximal AL & 2.08\,(3.04) & 11.32\,(14.46) & 26.76\,(28.27) & 46.82\,(53.85) \\
\texttt{OSQP} & ADMM & 2.96\,(4.07) & 11.85\,(14.04) & 26.30\,(30.65) & 41.16\,(47.41) \\
\texttt{SCS} & conic ADMM & 4.05\,(5.19) & 13.99\,(18.63) & 27.85\,(29.68) & 59.20\,(71.13) \\
\texttt{Clarabel} & conic IP & 5.76\,(6.05) & 34.32\,(39.74) & 75.56\,(87.12) & 129.4\,(139.0) \\
\texttt{ECOS} & conic IP & 9.50\,(9.72) & 60.27\,(70.98) & 224.1\,(236.6) & 575.2\,(625.1) \\
\texttt{CVXOPT} & IP & 3.41\,(4.03) & 20.27\,(24.08) & 66.70\,(73.60) & 150.7\,(166.3) \\
\texttt{HiGHS} & IP/simplex & 7.13\,(9.20) & 34.86\,(37.88) & 94.27\,(100.4) & 197.1\,(215.4) \\
\bottomrule
\end{tabular}
\end{table}

%% file: results_text.tex
Tables~\ref{tab:timing-sparse} and~\ref{tab:timing-cond} and
Fig.~\ref{fig:horizon} give the Python-mediated timings and
Table~\ref{tab:native} the native C++ ones; Table~\ref{tab:accuracy} gives the
quality of the returned solutions, Table~\ref{tab:overhead} the per-call
overhead, Table~\ref{tab:wsctl} the warm-start behavior, and
Table~\ref{tab:proxqp} a failure mode discussed separately below. Seven
observations stand out.

\begin{figure}[tbp]
\centering
\includegraphics[width=0.86\textwidth]{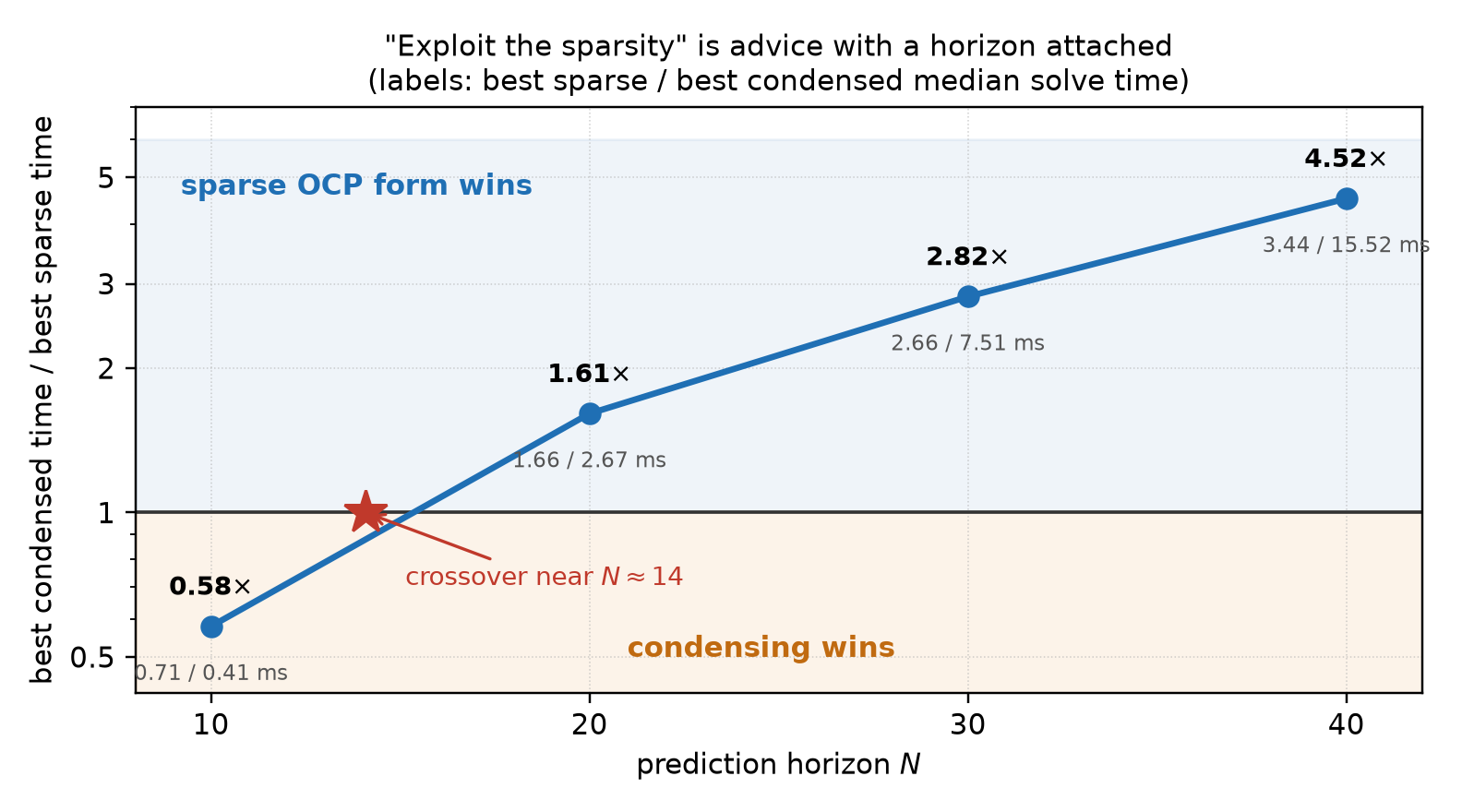}
\caption{Ratio of the best condensed solve time to the best sparse solve time, over
the horizon. Above the line the sparse OCP form wins, below it condensing
does; the two trade places at $N\approx14$ for this problem family. The small
labels give the two absolute medians being divided.}
\label{fig:crossover}
\end{figure}

\paragraph{1. Keeping the OCP sparsity pays off beyond a short horizon, but
not below it.}
On the sparse form the fastest solver is \texttt{PIQP} at every horizon, and
its cost grows almost linearly in $N$: $0.707$, $1.661$, $2.663$ and
$3.435$\,ms at $N=10,20,30,40$, i.e.\ a factor $4.9$ for a factor $4$ in
problem size. On the condensed form the fastest solver is \texttt{quadprog}
with $0.409$, $2.671$, $7.509$ and $15.517$\,ms, a factor $38$ over the same
range, as the dense Hessian grows quadratically. The two curves therefore
cross: at $N=10$ the condensed form is $1.7\times$ \emph{faster} than the best
sparse result, whereas at $N=40$ it is $4.5\times$ slower. For the
$30$\,ms control period assumed here, both formulations remain usable to
$N=40$, but only the sparse one leaves a comfortable margin. That condensing and sparsity exploitation trade places with horizon is not a
new phenomenon. It is the organizing axis of~\cite{kouzoupis2018}, which
measures the crossover and its dependence on the state dimension. What we add
is the location for a legged instance. For this problem family the crossover
sits between $N=10$ and $N=20$. We quote it as a range rather than a point
because we measured four horizons spaced ten apart; the $N\approx14$ in
Fig.~\ref{fig:crossover} is interpolation between two of them, not a
measurement.

\begin{figure}[tbp]
\centering
\includegraphics[width=0.95\textwidth]{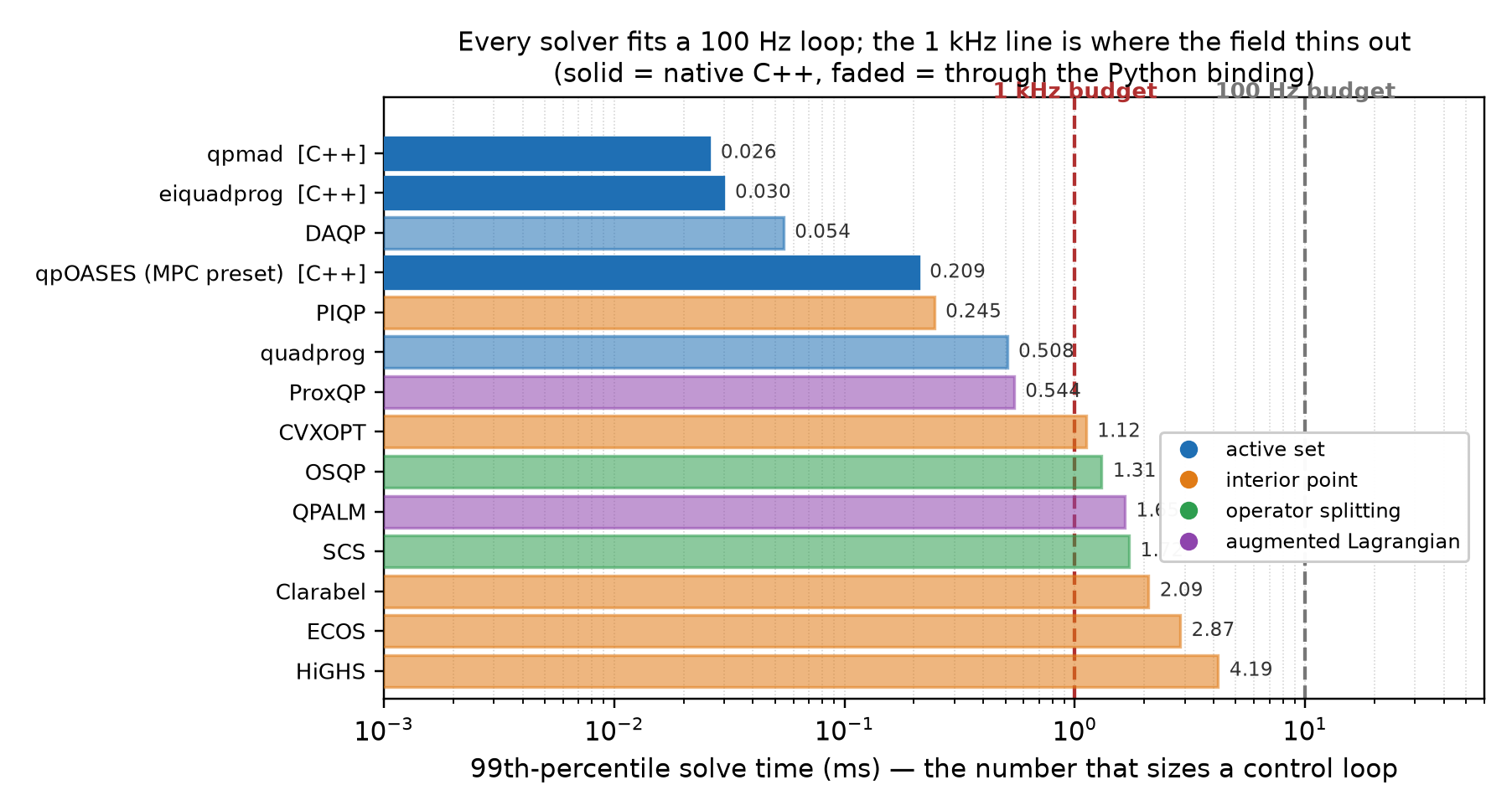}
\caption{Whole-body inverse dynamics: 99th-percentile solve time against two
common control-loop budgets. Every solver clears $100$\,Hz; the $1$\,kHz line
is where the field divides, and the four fastest are all active-set
methods. Solid bars are native C++, faded bars go through the
Python binding --- the difference between the two \texttt{qpOASES} entries and
everything else is discussed in observation~6.}
\label{fig:budget}
\end{figure}

\paragraph{2. Matching the solver to the structure matters more than picking a
``fast'' solver.}
\texttt{DAQP}, a dense dual active-set method, is the fastest solver in the
whole study on the small dense WBC problem ($0.029$\,ms median,
$0.054$\,ms at the 99th percentile, i.e.\ headroom for a $1$\,kHz loop by more
than an order of magnitude; Fig.~\ref{fig:budget} puts every solver against
that budget and the $30$\,ms MPC one) and is competitive on the condensed MPC
problems.
On the sparse form, where it must treat a $1000$-variable problem as dense, it
degrades to $107$\,ms, $31\times$ slower than \texttt{PIQP} on exactly the
same optimization problem. Conversely \texttt{Clarabel} and \texttt{PIQP},
strong on the sparse form, fall behind \texttt{quadprog} by $7$--$8\times$ once
the problem is condensed. No solver in Table~\ref{tab:timing-sparse} is
uniformly best, and the penalty for a structural mismatch is larger than the
spread between reasonable choices within the right structure.

\input{tab_accuracy}

\begin{figure}[tbp]
\centering
\includegraphics[width=0.99\textwidth]{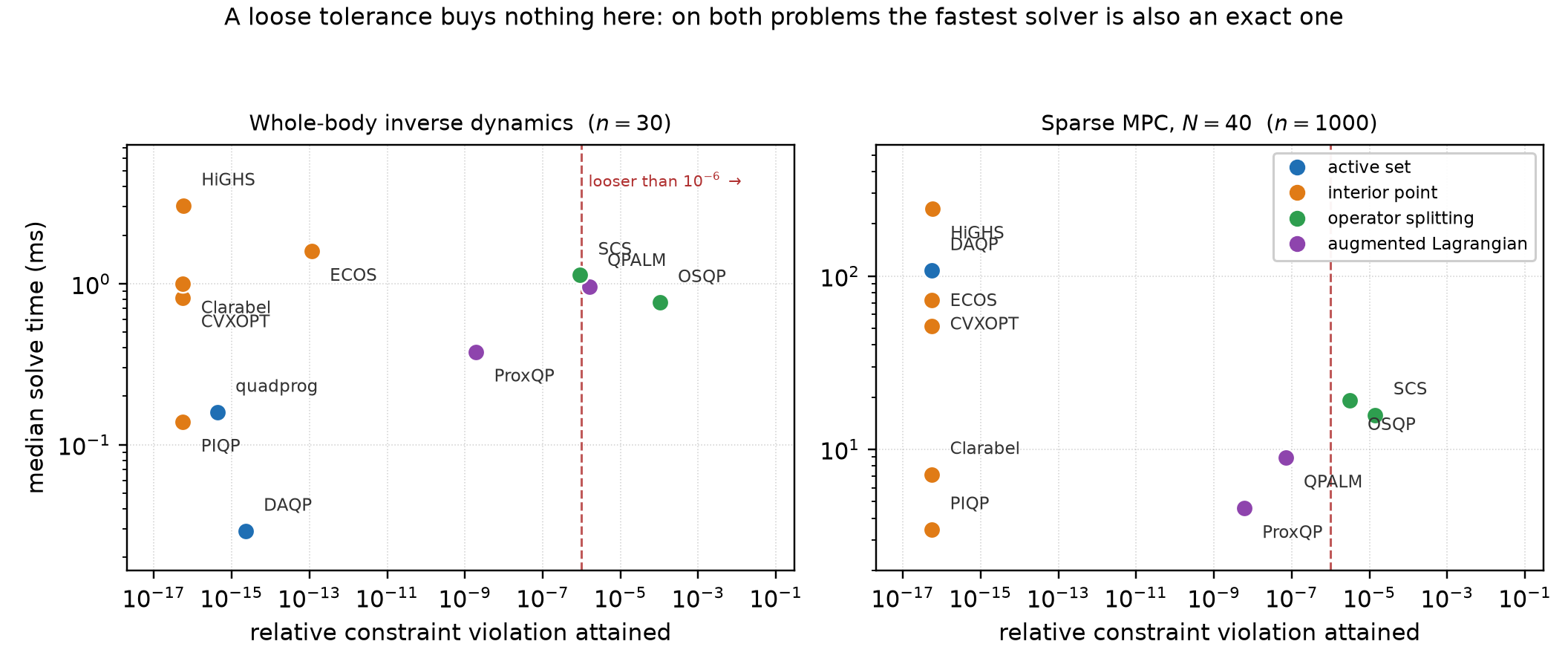}
\caption{Solve time against the inequality violation attained \emph{at each
solver's default settings}, colored by algorithm family. The families
separate cleanly. Two limits: the metric $\|\max(0,Cx-d)\|_\infty$ is
one-sided, so an interior-point method, which returns a strictly interior
point, scores zero by construction; and the horizontal axis shows where each
solver chose to stop, not what it can reach. Table~\ref{tab:matched} repeats
the comparison at a common tolerance with a two-sided measure.}
\label{fig:pareto}
\end{figure}

\paragraph{3. At default settings the solvers stop six orders of magnitude
apart; at a matched tolerance that becomes an ordinary speed--accuracy
trade-off.}
Table~\ref{tab:accuracy} and Fig.~\ref{fig:pareto} separate the solvers
sharply at their defaults. The interior-point and active-set solvers satisfy
every constraint to machine precision. Across the three instances their
relative inequality violation never exceeds $1.1\times10^{-13}$ and their
relative equality residual never exceeds $1.8\times10^{-12}$, while the
operator-splitting solvers stop at an inequality violation of up to
$1.0\times10^{-4}$ and the proximal augmented-Lagrangian ones at an equality
residual of up to $7.3\times10^{-4}$. That is not a defect:
\texttt{OSQP} ships $\epsilon_{\mathrm{abs}}=\epsilon_{\mathrm{rel}}=10^{-3}$,
\texttt{SCS} and \texttt{QPALM} $10^{-4}$, \texttt{ProxQP} $10^{-5}$, and each
delivers at or better than it documents.

What a defaults-only comparison cannot answer is what that accuracy costs.
We re-ran every solver at a common $\epsilon=10^{-9}$ wherever the interface
accepts one, and scored them by relative suboptimality
$|f(x)-f^\star|/\max(1,|f^\star|)$. The reference $f^\star$ is the mean of
the values returned by the independent high-accuracy solvers that agree on
each instance, so it belongs to no solver in the table; they agree to
$1.9\times10^{-11}$ on the sparse problem and $3.9\times10^{-11}$ on the WBC
one. Table~\ref{tab:matched} gives the result.

The interior-point and active-set solvers barely move, because they were
already there. The operator-splitting solvers pay for what they had been
skipping. On the WBC problem \texttt{OSQP} goes from $0.47$ to $4.04$\,ms, a
factor of $8.7$, and on the sparse $N=40$ problem from $18.5$ to
$56.4$\,ms, a factor of $3.1$. \texttt{QPALM} and \texttt{SCS} pay $2.2$ to
$3.2\times$ on the sparse problem and only $1.0$ to $1.3\times$ on the WBC
one, so the penalty is a property of the instance as much as of the solver.
\texttt{OSQP}'s suboptimality on WBC falls from $1.1\times10^{-3}$ to
$1.3\times10^{-9}$ once it is asked for it.

The trade-off itself is ordinary: operator splitting buys speed with
accuracy, as its own literature says. The size of the default gap is not. A
reader comparing default solve times is comparing solvers that stopped at
tolerances six orders of magnitude apart. Pick the tolerance from the units
of the constraint that binds, check the equality rows separately, and compare
like with like.

\input{tab_matched}

\paragraph{4. Most of what is called ``warm starting'' is setup reuse, and
one shipped default hides the rest.}
Measuring a warm start takes more care than it first appears, because two
distinct effects are easily conflated. We use one perturbation model for
every solver, on a closed-loop receding-horizon sequence: each QP is solved,
the first control block of its solution is applied to the single-rigid-body
dynamics, the state is propagated, the reference is shifted, and the
resulting sequence of linear terms is then replayed to every solver, so that
the loop is closed and all four still see exactly the same problems. Three
legs are timed separately: \emph{cold} (a fresh object each step),
\emph{reset} (one persistent object, the previous iterate discarded) and
\emph{warm} (the same object, iterate kept). Dividing cold by reset isolates
factorization and setup reuse; dividing reset by warm isolates the warm
start itself.

The reset leg is the whole difficulty, because it means something different
in each API. \texttt{OSQP} exposes a \texttt{warm\_start} flag.
\texttt{ProxQP} recomputes its guess every step unless
\texttt{initial\_guess} is switched to
\texttt{WARM\_START\_WITH\_PREVIOUS\_RESULT}, which must be done after the
first solve. \texttt{DAQP} retains its working set through \texttt{update()},
so a reset has to clear the active-set flags with a zeroed \texttt{sense}
array. \texttt{PIQP} offers no warm-start setting at all, so its warm leg
does not exist and we report none rather than a $1.00\times$ that would
present a missing facility as a measurement. Every leg is scored against the
exact optimum of each problem, so a leg that is fast because it stopped early
would show up as an accuracy gap instead of a speed-up.

\input{tab_wsctl}

\texttt{DAQP}'s large advantage is almost entirely \emph{setup reuse}:
$24.8\times$ from keeping the object, against $1.22\times$ from the warm
start on top, which sits at the harness's own $1.03\times$ drift floor. The
one unambiguous warm-start gain in the table belongs to \texttt{ProxQP}, at
$1.56\times$, and it is invisible unless \texttt{initial\_guess} is switched
explicitly. The familiar explanation, that an active set carries over between
neighboring problems while an interior-point iterate does not, is therefore
the wrong attribution for the largest number here. What a
persistent active-set solver mainly saves is the factorization it would
otherwise rebuild. The ratios hold to two decimal places across five
perturbation sizes spanning three orders of magnitude, from $0.014\%$ to
$10.4\%$ of $\|q\|$ per step, so none of them is an artifact of one step
size.

\texttt{OSQP}'s $1.00\times$ is the instructive case, because the solver does
have a warm start and this measurement cannot see it. OSQP tests its
termination criteria every \texttt{check\_termination} iterations, $25$ by
default, so no solve reports fewer than $25$ however good its starting point.
Table~\ref{tab:osqpct} repeats the reset-against-warm comparison at an
interval of $1$: the warm-started solve then converges in $1$ to $5$
iterations against $7$, and the speed-up appears, $5.6\times$ at the smallest
step and $1.3\times$ at the largest. What the default costs is visibility,
not iterations, and it is one line of configuration away.

\input{tab_osqpct}

The practical advice is unchanged and now better founded. Keep the solver
object alive, because that is where most of the gain is. Enable the warm
start explicitly, because it ships off in one of these solvers and masked in
another. And measure the interface that will be deployed. On this same sequence the
generic \texttt{initvals} argument returned between $0.94\times$ and
$1.34\times$, straddling the drift floor, for the three solvers
\texttt{qpsolvers} could reach in the measurement environment.

\paragraph{5. The binding overhead is a large share of a WBC solve.}
Before any optimization happens, the per-call floor already spans two orders
of magnitude between wrappers: $0.007$\,ms for \texttt{DAQP} against
$0.693$\,ms for \texttt{HiGHS} (Table~\ref{tab:overhead}). For the WBC problem that
floor is $36\%$ of \texttt{OSQP}'s total $0.763$\,ms and $23\%$ of
\texttt{HiGHS}'s $3.044$\,ms, while it is negligible for \texttt{DAQP}. Any
Python-mediated comparison of small dense QPs that does not report this floor
is partly ranking bindings rather than algorithms; Table~\ref{tab:native}
shows what disappears when the binding does.

\input{tab_native}

\paragraph{6. Native C++ timings are about three times faster, and one
default is catastrophic.}
Of the three dense C++ solvers run on the same data
(Table~\ref{tab:native}), \texttt{eiquadprog} is the fastest on the condensed
MPC problems
($0.187$ to $4.71$\,ms from $N=10$ to $N=40$), which is $2.2$ to $3.3\times$
faster than the best Python-mediated result in
Table~\ref{tab:timing-cond}; the gap is the binding overhead and the
interpreter, not the algorithms. \texttt{qpmad} is close behind and is the
fastest solver in the entire study on the WBC problem at $0.023$\,ms.

\begin{figure}[tbp]
\centering
\includegraphics[width=0.99\textwidth]{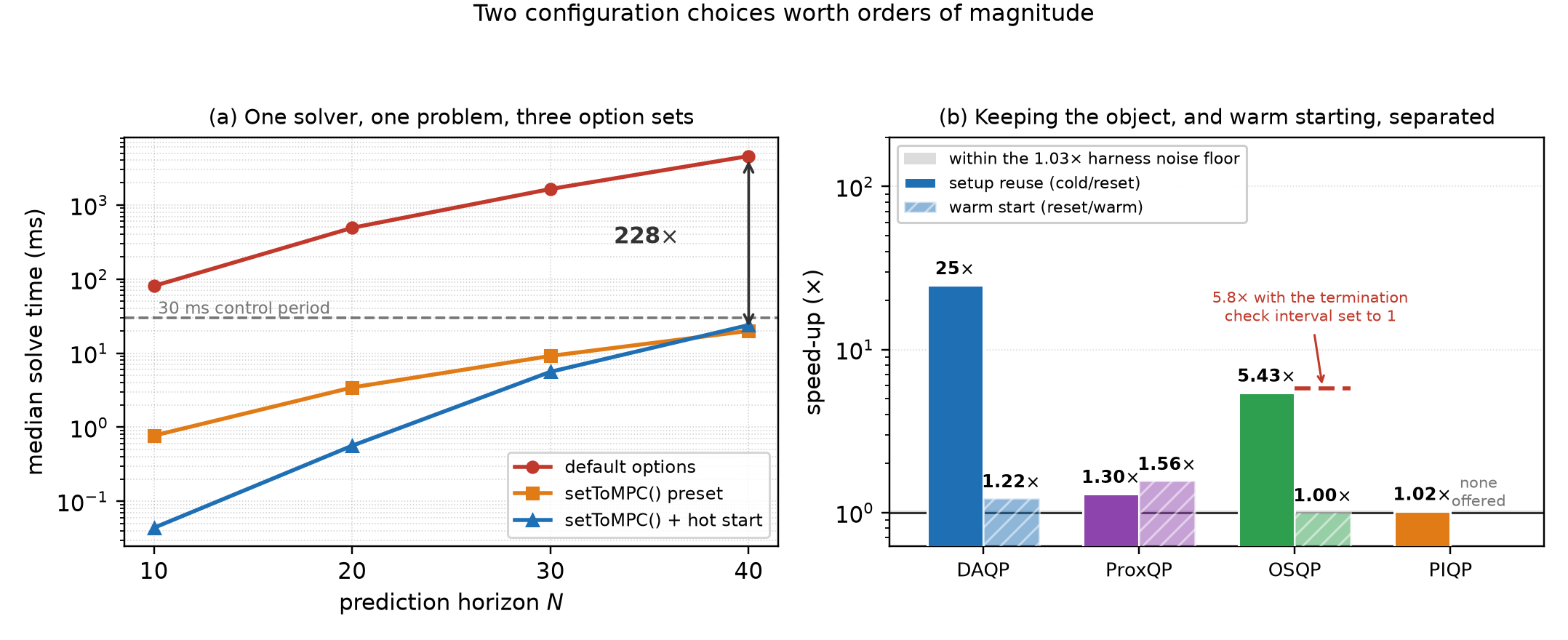}
\caption{Two configuration choices, each worth orders of magnitude.
\textbf{(a)} \texttt{qpOASES} on the condensed MPC problem under three option
sets: its defaults, its own \texttt{setToMPC()} preset, and that preset with a
parametric hot start. Only the lower two curves fit inside the $30$\,ms
control period at any horizon. \textbf{(b)} The two effects usually reported
as one warm-start number, separated on the sequence of
Table~\ref{tab:wsctl}: keeping the solver object alive (solid) and warm
starting on top of it (hatched). The shaded band is the harness's own drift
floor, below which a ratio cannot be told from no effect. \texttt{PIQP}
exposes no warm start to measure. \texttt{OSQP}'s appears only once its
default termination-check interval is lowered (dashed outline).}
\label{fig:defaults}
\end{figure}

\texttt{qpOASES} at its \emph{default} options is the outlier of the whole
study. Cold, it takes $80.7$\,ms at $N=10$ and $4.56$\,\emph{seconds} at
$N=40$, roughly $430$ and $970$ times slower than \texttt{eiquadprog} on the
identical problem. The library ships a preset for exactly this case, and
switching to \texttt{setToMPC()} cuts those to $0.77$\,ms and $20.0$\,ms, a
speed-up of $105\times$ to $228\times$ (Fig.~\ref{fig:defaults}a). The preset
changes ten options at once, among them the termination tolerance, so this is
not a single-knob comparison; what makes it a fair one here is that both
option sets return the same objective to eight digits at every horizon, at
relative violations below $10^{-14}$.
The optimal active set contains \emph{ten} of the $960$ constraints at
$N=40$, and \texttt{eiquadprog} finds that solution in $4.71$\,ms --- so the
$4.56$\,s is not the cost of a large problem but of a default option set
hunting for a handful of rows. The penalty also grows with size. On the WBC
problem the same default-versus-preset gap is only $9.6\times$. The same preset
is used, without comment, in the public dense benchmark of~\cite{qpmad_benchmark},
which reports it as a ``\textasciitilde50\% performance boost''; on these
problems it is worth two orders of magnitude more than that. A practitioner
who evaluates \texttt{qpOASES} at its defaults on a condensed MPC problem will
conclude that it is unusable, and the preset shows otherwise.

\paragraph{7. A real robot model moves the middle of the WBC ranking and
breaks nothing.}
Sect.~\ref{sec:protocol} is explicit that the WBC instance is structural
rather than analytic. To bound what that costs, we rebuilt the same QP from
the Unitree Go2 URDF with Pinocchio, so that $M$, $\eta$ and $J_c$ are the
robot's own, and timed both instances on one machine with one solver set.

What changes is the physics, not the conditioning. The synthetic optimum has
the feet carrying $21.7$\,N against a $117.7$\,N robot, three of the four on
the lower force bound; the Go2's carry $158.15$\,N against a weight of
$157.81$\,N, which is a stance. The conditioning is unremarkable on both
sides: $\mathrm{cond}(M)$ is $5.2$ for the synthetic instance and
$5.1\times10^{3}$ for the Go2, and the stacked contact Jacobian is full rank
in both cases, at condition number $6.2$ against $14.6$. A randomly
generated Jacobian is not pathological here, and neither is a real one.

Whether a change of ordering is a robot effect or just another draw of the
random instance is not something one seed can answer, so we repeated the
synthetic family across five seeds (\texttt{seed\_study.py}) and kept only
the moves that fall outside that variation. \texttt{DAQP} and
\texttt{quadprog} hold first and second on every seed and on the robot.
\texttt{ProxQP} and \texttt{PIQP} already exchange third and fourth places
between seeds of the synthetic instance, so their exchange on the robot is
not evidence of anything. What survives is a middle that genuinely reorders:
\texttt{SCS} rises from sixth or seventh to fifth, \texttt{QPALM} falls from
fifth to sixth, \texttt{Clarabel} rises from eighth or ninth to seventh,
\texttt{ECOS} from tenth or eleventh to ninth, and \texttt{CVXOPT} falls
furthest, from sixth-to-eighth on the synthetic instance to tenth on the
robot. Every solver that solves the synthetic instance also solves the robot
one. One solver cannot be placed at all: \texttt{OSQP} ranges from sixth to
eleventh across the five synthetic seeds, its median time varying by
$2.0\times$, so no statement about its position on this problem is
supported. The conclusion is narrower than one might expect
and worth stating plainly. For a dense QP of this size the ranking is driven
more by each solver's own overhead than by the instance, so a synthetic
instance of the right size and structure predicts the ordering at the
extremes but not in the middle. The released \texttt{wbc\_urdf.py} repeats
this on any model and posture, and reports $\mathrm{rank}(J_c)$,
$\mathrm{rank}(A_{\mathrm{eq}})$ and the total normal force, which is what
catches a contact set that is not the one intended.

\paragraph{A false infeasibility certificate at default settings.}
One result deserves separate emphasis because it is a correctness issue rather
than a performance one. \texttt{ProxQP} returns no solution on
\emph{every} condensed instance (Table~\ref{tab:timing-cond}). It is not
timing out. It declares the problem \texttt{PROXQP\_PRIMAL\_INFEASIBLE} after a
single iteration. The problem is feasible, and not marginally so. A two-line
analytic certificate settles it: putting every foot at a quarter of body
weight, $u_k=\mathbf{1}_4\otimes(0,0,\tfrac{1}{4}m|\gamma|)^\top
\in\mathbb{R}^{n_u}$ at every stage, gives
$\max(C_{\mathrm{fric}}u-h_{\mathrm{fric}})=-17.66$ against
$\|h_{\mathrm{fric}}\|_\infty=250$: a strictly interior point with a
$7.1\%$ margin, at every horizon. Slater's condition holds with room to
spare, so no primal-infeasibility certificate can exist, and no argument from
scaling or tolerance can rescue the status. Independently, \texttt{PIQP},
\texttt{DAQP} and \texttt{quadprog} return the same optimal value to six
decimals at every horizon (Table~\ref{tab:proxqp}), and lowering
\texttt{eps\_primal\_inf} from its default $10^{-4}$ to $10^{-6}$ makes
\texttt{ProxQP} converge in $8$--$10$ iterations to that same value.

We did not identify what triggers the test. Conditioning
is the obvious suspect, and the evidence is against it: \texttt{ProxQP}'s
primal-infeasibility test does not involve the Hessian; the sparse
formulation is worse conditioned and solves without incident; and the status,
the iteration count and the returned violation are flat across a two-order
range of condensed condition numbers. Identifying the trigger needs the
quantities inside the test at the firing iterate, which we did not
instrument.

In a control loop this is the dangerous kind of failure: a confident,
immediate, incorrect answer, which a supervisor will act on by triggering a
fallback. A formulation-level
remedy sits upstream of the whole question: relaxing the constraints that may
legitimately be violated into penalized slack variables, or replacing them
with relaxed barriers~\cite{zeilinger2014,grandia2023}, makes the online QP
feasible by construction, so no infeasibility status, correct or
incorrect, ever reaches the supervisor. The cost is that the constraint is no longer
hard and that the penalty weights become another set of parameters coupled to
the solver's conditioning. ProxQP is a well-regarded and
widely used solver, and on the sparse formulation it is the second fastest in
this study; the finding is specific to the condensed formulation at the
shipped threshold. It is reported against \texttt{proxsuite} $0.8.2$, as
recorded in the released \texttt{results.json}. The sweep of
Table~\ref{tab:proxqp} additionally sets \texttt{eps\_abs}$=10^{-9}$ and
\texttt{max\_iter}$=20000$ in every row, so we repeated it at genuinely
untouched defaults on a second platform, macOS on arm64, with a different
compiler and BLAS from the Linux host of Sect.~\ref{sec:protocol}, and the
behavior is identical: \texttt{PROXQP\_PRIMAL\_INFEASIBLE} at iteration~1 at
all four horizons, with the returned point violating the constraints by
$16.1$ to $18.6$\,N. Setting \texttt{eps\_primal\_inf}$=10^{-6}$ and changing
nothing else then yields \texttt{PROXQP\_SOLVED} in $5$--$7$ iterations at the
reference optimum.
\texttt{proxsuite} $0.8.3$ was released on 12 November 2026, after our
measurements, and its notes include ``\emph{Fix dual infeasibility detection
when some bounds are infinite}'', and our condensed problems are indeed
passed with an infinite lower bound on every inequality row. We therefore
examined that release before reporting this finding. The only functional
change between $0.8.2$ and $0.8.3$ is a single hunk in
\texttt{dense/utils.hpp} (mirrored in \texttt{sparse/utils.hpp}), and it lies
entirely inside \texttt{global\_dual\_residual\_infeasibility}: it corrects
how the \emph{dual} test treats rows whose bounds are infinite. The status we
observe is \texttt{PROXQP\_PRIMAL\_INFEASIBLE}, produced by the separate
primal test, which that release does not modify. The finding is therefore not
addressed by $0.8.3$. The solver's journal
description~\cite{proxqp2025} states a global convergence guarantee to the
closest feasible QP. What we observe is not a counterexample to that
guarantee but a property of the default primal-infeasibility
\emph{threshold}, which fires on the first iterate at every horizon, across
Hessian condition numbers from $2.7\times10^{5}$ at $N=10$ to
$3.0\times10^{7}$ at $N=40$.

%% file: tab_accuracy.tex
\begin{table}[tbp]
\centering
\scriptsize
\caption{Solution quality at each solver's \emph{default} settings, all three columns truncated to the power of ten at or below the value. \emph{ineq.} is the relative inequality violation $\|\max(0,Cx-d)\|_\infty/\max(1,\|d\|_\infty)$; \emph{eq.} is the relative equality residual $\|Ax-b\|_\infty/\max(1,\|b\|_\infty)$, shown as ``--'' where the instance has no equality constraints; \emph{gap} is the duality gap reported by \texttt{qpsolvers}. \xmark: no solution returned; \texttt{n/a}: the wrapper rejects sparse input.}
\label{tab:accuracy}
\setlength{\tabcolsep}{3pt}
\begin{tabular}{@{}l r r r r r r r r r@{}}
\toprule
& \multicolumn{3}{c}{\textbf{sparse $N=40$}} & \multicolumn{3}{c}{\textbf{condensed $N=40$}} & \multicolumn{3}{c}{\textbf{WBC}} \\
\textbf{Solver} & ineq. & eq. & gap & ineq. & eq. & gap & ineq. & eq. & gap \\
\midrule
\texttt{DAQP} & $10^{-17}$ & $10^{-15}$ & $10^{-14}$ & $10^{-15}$ & -- & $10^{-10}$ & $10^{-15}$ & $10^{-14}$ & $10^{-13}$ \\
\texttt{quadprog} & n/a & n/a & n/a & $10^{-18}$ & -- & $10^{-11}$ & $10^{-16}$ & $10^{-14}$ & $10^{-14}$ \\
\texttt{PIQP} & $0$ & $10^{-16}$ & $10^{-8}$ & $0$ & -- & $10^{-8}$ & $0$ & $10^{-15}$ & $10^{-11}$ \\
\texttt{ProxQP} & $10^{-9}$ & $10^{-7}$ & $10^{-5}$ & \xmark & \xmark & \xmark & $10^{-9}$ & $10^{-9}$ & $10^{-7}$ \\
\texttt{QPALM} & $10^{-8}$ & $10^{-4}$ & $10^{-3}$ & $10^{-7}$ & -- & $10^{-3}$ & $10^{-6}$ & $10^{-4}$ & $10^{-3}$ \\
\texttt{OSQP} & $10^{-6}$ & $10^{-4}$ & $10^{-1}$ & $10^{-5}$ & -- & $10^{-4}$ & $10^{-4}$ & $10^{-5}$ & $10^{-3}$ \\
\texttt{SCS} & $10^{-5}$ & $10^{-10}$ & $10^{-3}$ & $10^{-6}$ & -- & $10^{-5}$ & $10^{-7}$ & $10^{-9}$ & $10^{-5}$ \\
\texttt{Clarabel} & $0$ & $10^{-16}$ & $10^{-7}$ & $0$ & -- & $10^{-5}$ & $0$ & $10^{-15}$ & $10^{-9}$ \\
\texttt{ECOS} & $0$ & $10^{-13}$ & $10^{-5}$ & $0$ & -- & $10^{-2}$ & $10^{-13}$ & $10^{-12}$ & $10^{-6}$ \\
\texttt{CVXOPT} & $0$ & $10^{-15}$ & $10^{-5}$ & $0$ & -- & $10^{-3}$ & $0$ & $10^{-16}$ & $10^{-7}$ \\
\texttt{HiGHS} & $10^{-17}$ & $10^{-14}$ & $10^{-2}$ & $10^{-16}$ & -- & $10^{-2}$ & $10^{-17}$ & $10^{-14}$ & $10^{-5}$ \\
\bottomrule
\end{tabular}
\end{table}

%% file: tab_matched.tex
\begin{table}[tbp]
\centering
\scriptsize
\caption{Speed at each solver's default tolerance against speed at a common $\epsilon=10^{-9}$, with relative suboptimality $|f(x)-f^\star|/\max(1,|f^\star|)$, truncated to the power of ten at or below the value. The reference $f^\star$ is the mean of the values returned by the independent high-accuracy solvers that agree on each instance (sparse: \texttt{clarabel}, \texttt{daqp}, \texttt{piqp}, wbc: \texttt{clarabel}, \texttt{daqp}, \texttt{piqp}, \texttt{quadprog}), so it is not any one solver's own answer; their spread is sparse $1.9e-11$, wbc $3.9e-11$. ``A common $\epsilon$'' is not one setting: the per-solver mapping is listed in \texttt{matched\_tolerance.py} and recorded in the released JSON. \texttt{quadprog} terminates at an exact vertex and has no tolerance to set, so its two columns are identical by design, and it is absent from the sparse instance because its interface is dense-only. Measured on macOS/arm64 with \texttt{qpsolvers}~4.13.0.}
\label{tab:matched}
\setlength{\tabcolsep}{4pt}
\begin{tabular}{@{}l r r r r | r r r r@{}}
\toprule
& \multicolumn{4}{c|}{\textbf{sparse MPC, $N=40$}} & \multicolumn{4}{c}{\textbf{WBC}} \\
\textbf{Solver} & def.\ ms & $10^{-9}$ ms & sub@def & sub@$10^{-9}$ & def.\ ms & $10^{-9}$ ms & sub@def & sub@$10^{-9}$ \\
\midrule
\texttt{DAQP} & 57.93 & 55.88 & $10^{-11}$ & $10^{-11}$ & 0.02 & 0.02 & $10^{-11}$ & $10^{-11}$ \\
\texttt{quadprog} & -- & -- & -- & -- & 0.08 & 0.08 & $10^{-11}$ & $10^{-11}$ \\
\texttt{PIQP} & 2.27 & 2.26 & $10^{-11}$ & $10^{-11}$ & 0.10 & 0.10 & $10^{-11}$ & $10^{-11}$ \\
\texttt{ProxQP} & 1.14 & 1.37 & $10^{-7}$ & $10^{-9}$ & 0.12 & 0.12 & $10^{-8}$ & $10^{-10}$ \\
\texttt{QPALM} & 5.45 & 17.28 & $10^{-5}$ & $10^{-10}$ & 0.36 & 0.38 & $10^{-4}$ & $10^{-9}$ \\
\texttt{OSQP} & 18.46 & 56.43 & $10^{-5}$ & $10^{-10}$ & 0.46 & 4.04 & $10^{-3}$ & $10^{-9}$ \\
\texttt{SCS} & 9.80 & 21.65 & $10^{-6}$ & $10^{-11}$ & 0.48 & 0.63 & $10^{-6}$ & $10^{-11}$ \\
\texttt{Clarabel} & 5.07 & 5.35 & $10^{-10}$ & $10^{-12}$ & 0.60 & 0.60 & $10^{-11}$ & $10^{-11}$ \\
\texttt{ECOS} & 23.72 & 24.15 & $10^{-10}$ & $10^{-12}$ & 1.01 & 1.05 & $10^{-10}$ & $10^{-11}$ \\
\texttt{CVXOPT} & 29.04 & 36.96 & $10^{-7}$ & $10^{-11}$ & 0.55 & 0.58 & $10^{-8}$ & $10^{-10}$ \\
\texttt{HiGHS} & 225.5 & 225.2 & $10^{-9}$ & $10^{-9}$ & 0.89 & 0.88 & $10^{-11}$ & $10^{-11}$ \\
\bottomrule
\end{tabular}
\setlength{\tabcolsep}{6pt}
\end{table}

%% file: tab_wsctl.tex
\begin{table}[tbp]
\centering
\footnotesize
\caption{Warm starting decomposed, under one perturbation model for every solver. A closed-loop receding-horizon sequence of 60 condensed MPC problems ($N=10$): each QP is solved, its first control block is applied to the dynamics, the state is propagated and the reference shifted, and the resulting sequence is replayed to every solver. \emph{cold}: a fresh solver object each step. \emph{reset}: one persistent object, data updated, previous iterate discarded. \emph{warm}: the same, iterate kept. cold/reset is setup and factorization reuse; reset/warm is the warm start itself. The two right-hand columns give the range of each ratio over five perturbation sizes spanning 0.014\% to 10.4\% of $\|q\|$ per step; the decomposition is stable across them. \texttt{PIQP} exposes no warm-start setting, so its warm leg does not exist and is not reported as $1.00\times$. The harness's own drift floor is $1.03\times$, so a ratio below that is not distinguishable from no effect. Measured at a median step of 0.597\% on macOS/arm64.}
\label{tab:wsctl}
\begin{tabular}{@{}l r r r r r c c@{}}
\toprule
\textbf{Solver} & \textbf{cold} & \textbf{reset} & \textbf{warm} & \textbf{setup} & \textbf{warm} & \textbf{setup} & \textbf{warm} \\
 & (ms) & (ms) & (ms) & \textbf{reuse} & \textbf{start} & \textbf{range} & \textbf{range} \\
\midrule
\texttt{DAQP} & 0.268 & 0.011 & 0.009 & $24.8\times$ & $\mathbf{1.22\times}$ & 24.76--25.02 & 1.20--1.22 \\
\texttt{ProxQP} & 0.138 & 0.106 & 0.068 & $1.3\times$ & $\mathbf{1.56\times}$ & 1.29--1.30 & 1.56--1.57 \\
\texttt{OSQP} & 1.332 & 0.245 & 0.245 & $5.4\times$ & $\mathbf{1.00\times}$ & 5.39--5.43 & 1.00--1.00 \\
\texttt{PIQP} & 1.623 & 1.591 & -- & $1.0\times$ & -- & 1.02--1.02 & -- \\
\bottomrule
\end{tabular}
\end{table}

%% file: tab_osqpct.tex
\begin{table}[tbp]
\centering
\scriptsize
\caption{\texttt{OSQP}'s warm start, hidden and revealed. The solver checks its termination criteria every \texttt{check\_termination} iterations, $25$ by default, so no solve can report fewer than that however good its starting point is. At the default interval the reset and warm legs are indistinguishable at every perturbation size. At an interval of $1$ the same comparison shows the warm start converging in $1$ to $5$ iterations against $9$, and the speed-up appears. The effect is a reporting granularity, not an algorithmic limit, and it is one line of configuration away.}
\label{tab:osqpct}
\setlength{\tabcolsep}{2.5pt}
\begin{tabular}{@{}r rr rr c rr rr c@{}}
\toprule
 & \multicolumn{5}{c}{\textbf{\texttt{check\_termination}=25 (default)}} & \multicolumn{5}{c}{\textbf{=1}} \\
\cmidrule(lr){2-6}\cmidrule(lr){7-11}
\textbf{step} & \multicolumn{2}{c}{reset} & \multicolumn{2}{c}{warm} & \textbf{ratio} & \multicolumn{2}{c}{reset} & \multicolumn{2}{c}{warm} & \textbf{ratio} \\
(\%) & it & ms & it & ms & & it & ms & it & ms & \\
\midrule
0.014 & 25 & 0.236 & 25 & 0.235 & $1.00\times$ & 7 & 0.124 & 1 & 0.022 & $\mathbf{5.56\times}$ \\
0.126 & 25 & 0.236 & 25 & 0.235 & $1.00\times$ & 7 & 0.140 & 1 & 0.022 & $\mathbf{6.49\times}$ \\
0.597 & 25 & 0.235 & 25 & 0.235 & $1.00\times$ & 7 & 0.124 & 1 & 0.022 & $\mathbf{5.75\times}$ \\
2.764 & 25 & 0.236 & 25 & 0.235 & $1.01\times$ & 7 & 0.124 & 2 & 0.042 & $\mathbf{2.95\times}$ \\
10.405 & 25 & 0.235 & 25 & 0.235 & $1.00\times$ & 7 & 0.124 & 5 & 0.093 & $\mathbf{1.33\times}$ \\
\bottomrule
\end{tabular}
\setlength{\tabcolsep}{6pt}
\end{table}

%% file: tab_native.tex
\begin{table}[tbp]
\centering
\scriptsize
\caption{Native C++ timings, in milliseconds (99th percentile in brackets), for the three dense solvers that have no ready Python binding. These numbers cross no language boundary and so carry none of the per-call overhead of Table~\ref{tab:overhead}. \emph{cold} builds a fresh solver object for every solve; \emph{hot} keeps one object and re-solves after perturbing the linear term, which is the parametric warm start qpOASES is designed for; \emph{MPC} means the solver's own \texttt{setToMPC()} option preset instead of its defaults. On the \emph{cold} and \emph{MPC} rows all solvers agree with the Python reference optimum of Table~\ref{tab:proxqp} to the digits shown there, which is the correctness check on this harness. The \emph{hot} rows re-solve after perturbing the linear term and therefore return the optimum of a different problem, differing in the fourth to sixth significant figure.}
\label{tab:native}
\setlength{\tabcolsep}{2.5pt}
\begin{tabular}{@{}l r r r r r@{}}
\toprule
\textbf{Solver} & \textbf{$N=10$} & \textbf{$N=20$} & \textbf{$N=30$} & \textbf{$N=40$} & \textbf{WBC} \\
\midrule
\texttt{qpmad} & 0.260\,(0.291) & 1.15\,(1.21) & 2.59\,(2.63) & 5.23\,(5.36) & 0.023\,(0.026) \\
\texttt{eiquadprog} & 0.187\,(0.216) & 0.918\,(0.976) & 2.29\,(2.34) & 4.71\,(4.88) & 0.028\,(0.030) \\
\texttt{qpOASES} default, cold & 80.68\,(125.0) & 490.8\,(1123.4) & 1638.7\,(1858.6) & 4563.9\,(7088.1) & 1.79\,(1.86) \\
\texttt{qpOASES} MPC, cold & 0.772\,(0.866) & 3.44\,(3.53) & 9.17\,(9.43) & 19.98\,(20.57) & 0.186\,(0.209) \\
\texttt{qpOASES} default, hot & 0.056\,(0.224) & 1.01\,(2.37) & 13.60\,(22.99) & 83.06\,(118.9) & 0.004\,(0.005) \\
\texttt{qpOASES} MPC, hot & 0.044\,(0.122) & 0.564\,(1.08) & 5.60\,(8.96) & 23.91\,(33.82) & 0.003\,(0.003) \\
\bottomrule
\end{tabular}
\end{table}

%% file: tab_overhead.tex
\begin{table}[tbp]
\centering
\footnotesize
\caption{Per-call overhead floor: the median time to solve a two-variable QP, i.e. the fixed cost of one call through the Python binding. It is included in every timing in Tables~\ref{tab:timing-sparse} and~\ref{tab:timing-cond}, and it is what Table~\ref{tab:native} avoids by staying in C++. For the WBC problem this floor is $36\%$ of \texttt{OSQP}'s total solve time and $23\%$ of \texttt{HiGHS}'s.}
\label{tab:overhead}
\begin{tabular}{@{}l r l r@{}}
\toprule
\textbf{Solver} & \textbf{Overhead (ms)} & \textbf{Solver} & \textbf{Overhead (ms)} \\
\midrule
\texttt{DAQP} & 0.007 & \texttt{SCS} & 0.128 \\
\texttt{quadprog} & 0.009 & \texttt{Clarabel} & 0.126 \\
\texttt{PIQP} & 0.012 & \texttt{ECOS} & 0.081 \\
\texttt{ProxQP} & 0.018 & \texttt{CVXOPT} & 0.238 \\
\texttt{QPALM} & 0.118 & \texttt{HiGHS} & 0.693 \\
\texttt{OSQP} & 0.278 &  &  \\
\bottomrule
\end{tabular}
\end{table}

%% file: tab_proxqp.tex
\begin{table}[tbp]
\centering
\footnotesize
\caption{ProxQP on the \emph{condensed} MPC problem of Eq.~\eqref{eq:mpc}, as its primal-infeasibility threshold \texttt{eps\_primal\_inf} is varied. At the shipped default the solver declares the problem primal infeasible after a single iteration, at every horizon; the problem is in fact feasible, and PIQP and DAQP agree on the optimal value to nine or more significant digits. Each cell gives the returned status and the iteration count.}
\label{tab:proxqp}
\begin{tabular}{@{}l c c c c@{}}
\toprule
\texttt{eps\_primal\_inf} & $N=10$ & $N=20$ & $N=30$ & $N=40$ \\
\midrule
$10^{-4}$ (default) & \xmark\,infeas. (1) & \xmark\,infeas. (1) & \xmark\,infeas. (1) & \xmark\,infeas. (1) \\
$10^{-5}$ & \xmark\,infeas. (4) & \xmark\,infeas. (4) & \xmark\,infeas. (4) & \xmark\,infeas. (2) \\
$10^{-6}$ & \cmark\,solved (8) & \cmark\,solved (9) & \cmark\,solved (10) & \cmark\,solved (10) \\
$10^{-8}$ & \cmark\,solved (8) & \cmark\,solved (9) & \cmark\,solved (10) & \cmark\,solved (10) \\
$0$ & \cmark\,solved (8) & \cmark\,solved (9) & \cmark\,solved (10) & \cmark\,solved (10) \\
\midrule
\multicolumn{5}{@{}l}{\emph{reference optimal value from independent solvers}}\\
\texttt{PIQP} & -26.448097 & -431.925708 & -2754.110348 & -10862.101803 \\
\texttt{DAQP} & -26.448097 & -431.925708 & -2754.110348 & -10862.101803 \\
\texttt{ProxQP} ($10^{-6}$) & -26.448097 & -431.925708 & -2754.110348 & -10862.101803 \\
\bottomrule
\end{tabular}
\end{table}

%% file: gpu.tex
\section{GPU Acceleration: Where It Helps, and Where It Cannot}
\label{sec:gpu}

Every solver measured so far runs on a CPU core. Legged robots increasingly
carry an embedded GPU, the Jetson Orin NX of Sect.~\ref{sec:public} among
them, and a growing literature asks whether it should solve the QP. For
legged control the answer turns on a quantity that literature rarely reports:
the fixed cost of using the device at all. We review the work and measure
that cost.

\subsection{What has been built}

\paragraph{GPU QP solvers.}
The most direct approach ports a QP algorithm onto the device.
\texttt{cuOSQP}~\cite{schubiger2020gpu} replaces OSQP's direct
$LDL^\top$ factorization with a GPU conjugate-gradient solve and moves all
vector and matrix work onto the device, reporting speed-ups ``up to two
orders of magnitude'' over the CPU implementation on large problems.
\texttt{ReLU-QP}~\cite{bishop2024reluqp} takes a more surprising route: it
reformulates the same ADMM iteration \emph{exactly} as a deep, weight-tied
neural network with ReLU activations, so the solve becomes a sequence of
dense matrix products and clamps that any machine-learning runtime executes
on a GPU. It is the one GPU QP solver benchmarked directly on legged
whole-body MPC, and its numbers are worth quoting: on an Atlas balancing
task it reaches $1348$--$2599$\,Hz against $65$--$206$\,Hz for OSQP and
$66$--$192$\,Hz for ProxQP, a $10$ to $20\times$ gain. On a quadruped with a
6-DoF arm the same solver runs at $834$\,Hz against OSQP's $325$\,Hz --- only
$2.6\times$, which the authors attribute to ``the very large number of
constraints in the problem and ReLU-QP's current inability to leverage matrix
sparsity''. Only four of the fourteen solvers measured in Sect.~\ref{sec:own-bench} have
a GPU counterpart at all: \texttt{OSQP}, \texttt{SCS}, \texttt{Clarabel} and
\texttt{PIQP}. \texttt{PIQP}'s counterpart, the batched and differentiable
\texttt{cuPIQP}~\cite{cupiqp}, is as yet unpublished. Ten have none:
\texttt{qpOASES}, \texttt{DAQP}, \texttt{ProxQP}, \texttt{QPALM},
\texttt{qpmad}, \texttt{eiquadprog}, \texttt{quadprog}, \texttt{CVXOPT},
\texttt{ECOS} and \texttt{HiGHS}, and neither does \texttt{HPIPM}. GPU work in legged control has gone instead into
solver-native designs and into sampling-based methods, which is the more
useful way to organize it than solver by solver.

\paragraph{GPU trajectory optimization.}
A second line keeps the optimal-control structure and parallelizes the linear
algebra inside it. \texttt{MPCGPU}~\cite{adabag2024mpcgpu} solves the Schur
complement of the trajectory-optimization KKT system with a GPU
preconditioned conjugate-gradient method, reporting that the linear solve
beats CPU alternatives ``by at least 10x for a majority of solves and 3.6x on
average'' and reaching kilohertz rates with horizons up to $512$ knot points
--- on a 7-DoF manipulator, not a legged robot. The closest legged
equivalent is \texttt{MPX}~\cite{amatucci2026mpx}, a JAX primal-dual iLQR
using parallel associative scans, demonstrated on an Aliengo quadruped and a
Talos humanoid, with ``up to a 60\% improvement in runtime for Whole Body
Dynamics (WB)MPC and a 700\% improvement for Single Rigid Body Dynamics
(SRBD)-MPC'' over \texttt{acados} and \texttt{crocoddyl}. Most recently,
GPU-SLS~\cite{fang2026gpusls} closes a $50$\,Hz whole-body loop on Unitree Go2
hardware ($61$ states) with a GPU ADMM QP backend. Underneath all of these,
\texttt{GRiD}~\cite{plancher2022grid} supplies GPU rigid-body dynamics and
analytical gradients, the other half of the cost of a nonlinear MPC
iteration.

\paragraph{GPU-parallel symbolic evaluation.}
\texttt{CusADi}~\cite{jeon2025cusadi} compiles arbitrary CasADi expressions
into CUDA kernels callable from PyTorch, so thousands of instances of the same
symbolic function evaluate at once; the authors demonstrate MPC for the MIT
Humanoid across thousands of IsaacGym environments and report ``a ten-fold
speedup relative to similar MPC implementation on the CPU''. This is a
different use of a GPU from the previous paragraph, and the distinction
matters: the win is in \emph{how many} problems are solved, not how fast one
of them is.

\paragraph{Sampling-based MPC, which solves no QP at all.}
A thread that dominates recent GPU locomotion work sits outside this paper's
scope but should be named, because it is easy to conflate with GPU QP solving.
Controllers in the MPPI family~\cite{williams2017mppi} evaluate thousands of
randomly perturbed control sequences in parallel and take a weighted average;
they solve no QP, require no convexity, and map onto a GPU almost perfectly.
Recent work applies them to quadrupeds at torque level with real-time rates
\cite{xue2025dialmpc}. Choosing between this and a QP is not choosing between
two solvers but between two formulations with different guarantees: a QP
returns a constrained optimum, a sampling controller returns the best of what
it happened to try.

\subsection{The fixed cost of using the device}
\label{sec:gpu-floor}

The scale dependence in \texttt{ReLU-QP}'s own results is the crux for legged
control, because the WBC problem of Eq.~\eqref{eq:wbc} is \emph{small}: it
solves in $0.023$\,ms natively on one CPU core (Table~\ref{tab:native}).
Before a GPU solver can improve on that, it has to pay for the round trip.
Table~\ref{tab:gpu} measures what that costs on the test machine's GPU.

\input{tab_gpu}

An empty kernel launch followed by a synchronize costs $0.010$\,ms, which is
$43\%$ of the entire CPU solve. Moving the problem data to the device and the
answer back costs $0.027$\,ms --- more than solving the whole QP on the CPU,
before a single arithmetic operation. And the smallest useful step inside a
dense QP iteration, a $30\times30$ Cholesky factorization, takes $0.050$\,ms
on the GPU against $0.0056$\,ms on one CPU core.

Both numbers are measured through CuPy, so both carry Python dispatch: the
GPU figures include a fresh device allocation per call, and a large share of
the \texttt{numpy} baseline is dispatch too. The two biases cut in opposite
directions. A native CUDA implementation with pinned memory and a persistent
stream would beat the GPU figures, which narrows the margin we rely on; the
inflated CPU baseline flatters the batch crossover. We therefore argue from
the kernel-launch cost and the shape of the batch curve, not from the round
trip alone. On an integrated-memory device such as the Jetson Orin NX, the
host-to-device copy disappears, so the round-trip figure does not transfer
there.

The batched half of Table~\ref{tab:gpu} shows where this reverses. Solving
one factorization at a time, the GPU reaches $0.08\times$ the CPU's
throughput; at $B=8$ it is still behind at $0.63\times$; by $B=64$ it is
ahead at $3.9\times$, and at $B=512$ at $9.1\times$. The crossover on this
hardware sits between $8$ and $64$ simultaneous problems. A legged robot's
control loop presents the device with exactly one.

\subsection{Reading the two threads together}

Our measurement and the GPU-MPC literature are not in conflict; they describe
different regimes, and the GPU authors draw the line themselves. The authors
of \texttt{cuOSQP} conclude that ``GPUs are not suited for solving small
problems for which the CPU implementation is generally much
faster''~\cite{schubiger2020gpu}, because ``a kernel launch and data transfer
come with a constant overhead that cannot be amortized''. \texttt{GRiD}'s
authors find that ``the CPU is still faster than the GPU for all individual
computations, showing that GPU acceleration only makes sense when there is
sufficient parallel work to be done''~\cite{plancher2022grid}.
\texttt{MPCGPU} is ``only marginally slower at the smallest problem
size''~\cite{adabag2024mpcgpu}. Most pointedly, \texttt{MPX}'s legged whole-body results show
\texttt{crocoddyl} winning below $80$ nodes. Its authors attribute that to
``overhead involved in transferring data to and from the GPU, which, for
smaller problems, can be detrimental'' --- a round trip they
measure at ``approximately 1ms''~\cite{amatucci2026mpx}, some $40$ times our
WBC solve. And a legged hardware paper that evaluated the GPU option rejected it
deliberately. It found that MJX and Isaac can simulate far more parallel
rollouts, but that ``the individual rollout speeds are insufficient to
close the real-time feedback loop'', the authors parallelized on a multi-core
CPU instead~\cite{alvarezpadilla2025wbmppi}.

The resulting picture is consistent, and it happens to divide along the same
seam as Sect.~\ref{sec:two-families}.

\textbf{Where the GPU wins.} Long horizons and large state dimensions, where one solve is itself big
enough to fill the device. And batch workloads, where the question is
throughput rather than latency --- parameter sweeps, MPC
inside a reinforcement-learning loop, massively parallel simulation, and
sampling-based control. The MPC layer of a legged stack can cross into the
first regime at long horizons, and \texttt{MPX} and GPU-SLS show it happening
on real quadrupeds.

\textbf{Where it cannot.} A single small dense QP on the critical path of a
$1$\,kHz loop. The WBC problem is below the size at which a GPU has anything
to parallelize, and its per-call overhead alone exceeds the entire CPU solve.
No amount of device throughput recovers a latency floor. This is not a
criticism of GPU solvers --- it is a statement about where the boundary lies,
and for a legged robot that boundary runs between the two QP templates of
Sect.~\ref{sec:two-families}. We state it as a conclusion about discrete
GPUs measured through a Python array library; confirming it would need a
native CUDA probe on an integrated-memory embedded device, which we did not
have.

%% file: tab_gpu.tex
\begin{table}[tbp]
\centering
\footnotesize
\caption{What a GPU costs before it computes anything, measured on NVIDIA GeForce RTX 4050 Laptop GPU with CuPy~14.2.0. These are measured through a Python array library and are therefore an \emph{upper} bound on a native implementation's fixed cost; the CPU baseline is likewise a \texttt{numpy} call of which a large share is dispatch. In the upper block the ratio is against the whole-body QP of Eq.~\eqref{eq:wbc}, which \texttt{qpmad} solves natively on one CPU core in 0.0230\,ms (Table~\ref{tab:native}); a $30\times30$ Cholesky is the smallest useful linear-algebra step inside one iteration of a dense QP solver of that size. In the lower block the ratio is the GPU's throughput on a batch of $B$ such factorizations against the same $B$ performed one after another on one CPU core.}
\label{tab:gpu}
\setlength{\tabcolsep}{5pt}
\begin{tabular}{@{}p{7.0cm} r r@{}}
\toprule
\textbf{Operation} & \textbf{median (ms)} & \textbf{vs.\ WBC} \\
\midrule
empty kernel launch $+$ synchronize & 0.0100 & $0.44\times$ \\
host\,$\to$\,device\,$\to$\,host round trip of the problem data (23808\,B) & 0.0269 & $1.17\times$ \\
one $30\times30$ Cholesky, on the GPU & 0.0499 & $2.17\times$ \\
one $30\times30$ Cholesky, on one CPU core & 0.0056 & $0.24\times$ \\
\midrule
\emph{batched $30\times30$ Cholesky} & \textbf{per problem (ms)} & \textbf{speed-up} \\
\midrule
$B=1$ & 0.07157 & $0.08\times$ \\
$B=8$ & 0.00898 & $0.63\times$ \\
$B=64$ & 0.00143 & $3.94\times$ \\
$B=512$ & 0.00062 & $9.06\times$ \\
$B=4096$ & 0.00068 & $8.32\times$ \\
\bottomrule
\end{tabular}
\setlength{\tabcolsep}{6pt}
\end{table}

%% file: guide.tex
\section{Practical Guide for Solver Selection}
\label{sec:guide}

Combining the published evidence of Sect.~\ref{sec:public} with the
measurements of Sect.~\ref{sec:own-bench} gives the decision procedure of
Fig.~\ref{fig:flow}. It is deliberately phrased around properties of the
\emph{problem}, because those are known before any solver is chosen, and
every branch carries the measured number that justifies it.

\begin{figure}[tbp]
\centering
\footnotesize
\resizebox{\textwidth}{!}{%
\begin{tikzpicture}[
  box/.style   = {draw, rounded corners=2pt, align=center, inner sep=4pt,
                  font=\scriptsize, text width=#1},
  dec/.style   = {draw, diamond, aspect=3.1, align=center, inner sep=-1pt,
                  font=\scriptsize, fill=blue!5, text width=27mm},
  pick/.style  = {box=46mm, fill=green!8, draw=green!45!black},
  lbl/.style   = {font=\scriptsize\itshape, inner sep=2pt},
  ar/.style    = {-{Latex[length=1.8mm]}, thick}
]
\node[box=34mm, fill=gray!10]                 (start)   {your online QP};
\node[dec, below=7mm of start]                (dense)   {small and dense?\\[-2pt]
       \tiny WBC / inverse dynamics, $n\lesssim100$};
\node[dec, below=13mm of dense]               (horizon) {is the MPC horizon\\[-2pt]$N\lesssim14$?};
\node[coordinate, below=13mm of horizon]      (j) {};

\node[pick, right=16mm of dense]   (wbc)
     {\textbf{dense dual active set}\\ \texttt{qpmad} $\cdot$ \texttt{eiquadprog}
      $\cdot$ \texttt{DAQP}\\[2pt]
      \tiny $0.026$\,ms native C++ to $0.054$\,ms Python-mediated, 99th
      percentile, at machine precision};
\node[pick, right=16mm of horizon] (short)
     {\textbf{condense, then dense active set}\\ \texttt{qpOASES}
      (\texttt{setToMPC}) $\cdot$ \texttt{quadprog}\\[2pt]
      \tiny $1.7\times$ faster than the sparse form at $N=10$};
\node[pick, right=16mm of j]       (long)
     {\textbf{keep the OCP sparsity}\\ \texttt{PIQP} $\cdot$ \texttt{Clarabel}
      $\cdot$ \texttt{HPIPM}\\[2pt]
      \tiny $4.5\times$ faster than condensing at $N=40$;\\
      at long horizons a GPU solver becomes worth testing (Sect.~\ref{sec:gpu})};

\node[box=132mm, fill=red!4, draw=red!45!black, below=16mm of j.south,
      anchor=north, xshift=28mm] (checks)
     {\textbf{Whichever branch you took, three checks before you ship.}\\[1pt]
      \tiny Each is a failure we measured in Sect.~\ref{sec:own-bench}, with a
      well-regarded solver at its shipped settings.\\[4pt]
      \scriptsize
      \textbf{1 \ Defaults.}\ Solve one instance with two independent solvers
      and compare the optima.\\
      \tiny A default threshold made ProxQP call a feasible problem infeasible;
      default options made qpOASES $228\times$ slower than its own MPC preset.\\[3pt]
      \scriptsize \textbf{2 \ Tolerance.}\ Read the residual in the units of the
      constraint that binds, not in the solver's defaults.\\
      \tiny On the WBC problem OSQP stopped at a relative violation of
      $1\!\times\!10^{-4}$, against DAQP's $2\!\times\!10^{-15}$.\\[3pt]
      \scriptsize \textbf{3 \ Warm start.}\ If you chose the solver for it, use
      its stateful API --- and measure it.\\
      \tiny Keeping the solver object alive was worth $24.8\times$; the warm
      start on top of that, $1.6\times$.};

\draw[ar] (start)   -- (dense);
\draw[ar] (dense)   -- node[lbl, above] {yes} (wbc);
\draw[ar] (dense)   -- node[lbl, right] {no}  (horizon);
\draw[ar] (horizon) -- node[lbl, above] {yes} (short);
\draw[ar] (horizon) -- node[lbl, right] {no}  (j) -- (long);
\draw[ar] (long.south) -- ([xshift=18mm]checks.north);
\end{tikzpicture}}
\caption{Solver selection for a legged-robot stack, with the measurement
behind each branch. The three checks at the bottom are not generic advice:
each one is a failure mode that occurred in Sect.~\ref{sec:own-bench} with a
well-regarded solver at its shipped settings.}
\label{fig:flow}
\end{figure}

\subsection{Three cases the flowchart does not cover}

\paragraph{Fixed formulation on a microcontroller.}
When the dimensions and the sparsity pattern never change and the target is
an embedded core, code generation (\texttt{CVXGEN}, \texttt{CVXPYgen},
\texttt{FORCESPRO}) buys a tight footprint and predictable timing, at the cost
of regenerating whenever the formulation changes.

\paragraph{A GPU in the loop.}
Worth testing for the MPC layer at long horizons, and worth using for batched
work --- MPC inside a learning loop, parameter sweeps, sampling-based control.
Not worth testing for the WBC layer: Sect.~\ref{sec:gpu-floor} measures a
device round trip at $0.027$\,ms against a $0.023$\,ms CPU solve, and the
device needs tens of simultaneous problems before it breaks even.

\paragraph{Constraints beyond a QP.}
If the pipeline needs second-order cones or certificates, \texttt{Clarabel}
covers QPs as a special case; expect overhead relative to a QP-specialized
solver in the tightest loops --- on our condensed problems it was $8\times$
behind the best dense solver.

\subsection{What the measurements say that a solver ranking does not}

What follows generalizes beyond the particular solvers measured here.

\paragraph{The formulation outranks the solver.}
Giving one solver the same optimization problem in the structure it is built
for instead of the other structure was worth $32\times$ at $N=40$ for
\texttt{PIQP}, and across solvers the two structures spread the solve time by
$31\times$ on one and the same problem. The right structure is not fixed
either: Fig.~\ref{fig:crossover} shows the sparse and condensed forms trading
places at $N\approx14$. Decide the formulation first, then shop for a solver
inside it.

\paragraph{Defaults are a hypothesis about the problem, not a setting.}
A solver's defaults encode its authors' assumptions about scaling and
conditioning. A legged robot's QPs --- forces in newtons, accelerations in
$\mathrm{rad/s^2}$, a condensed Hessian with condition number in the hundreds
of thousands --- routinely fall outside those assumptions. Three of the results in this paper are default-value problems rather than
algorithm problems, and each is one line of configuration away from
disappearing.

\paragraph{Benchmark the interface you will deploy.}
The gap between a generic Python wrapper and a native C++ build was a factor
of $2.2$ to $3.3$ in raw solve time, and keeping one solver object alive
instead of rebuilding it was worth $24.8\times$ on top of that.
A ranking produced through one interface does not transfer to the other. If
the deployment is C++ with a persistent solver object, the benchmark has to be
too.

%% file: resub.bbl
\begin{thebibliography}{99}

\bibitem{stark2025benchmarking}
Stark, F., Middelberg, J., Mronga, D., Vyas, S., Kirchner, F.: Benchmarking Different QP Formulations and Solvers for Dynamic Quadrupedal Walking. In: \emph{Proc. IEEE Int. Conf. on Robotics and Automation (ICRA)}, pp. 14412--14418. IEEE (2025). \doi{10.1109/ICRA55743.2025.11128397}

\bibitem{Humanoid_rev}
Gu, Z., Li, J., Shen, W., Yu, W., Xie, Z., McCrory, S., Cheng, X., Shamsah, A., Griffin, R., Liu, C.K., Kheddar, A., Peng, X.B., Zhu, Y., Shi, G., Nguyen, Q., Cheng, G., Gao, H., Zhao, Y.: Humanoid Locomotion and Manipulation: Current Progress and Challenges in Control, Planning, and Learning. \emph{IEEE/ASME Transactions on Mechatronics} \textbf{31}(2), 2300--2330 (2026). \doi{10.1109/TMECH.2025.3579247}

\bibitem{van2024fusion}
Dinh, V.N., Phan, T.D., Park, C.H., Kim, G.-W.: Fusion Consistency for Industrial Robot Navigation: An Integrated SLAM Framework with Multiple 2D LiDAR-Visual-Inertial Sensors. \emph{Computers and Electrical Engineering} \textbf{120}, 109607 (2024). \doi{10.1016/j.compeleceng.2024.109607}

\bibitem{namlearning}
Dinh, V.N., Kim, G.-W.: Learning Observation Model for Factor Graph Based-State Estimation Using Intrinsic Sensors. \emph{IEEE Transactions on Automation Science and Engineering} \textbf{20}(3), 2049--2062 (2023). \doi{10.1109/TASE.2022.3193411}

\bibitem{dicarlo2018}
Di Carlo, J., Wensing, P.M., Katz, B., Bledt, G., Kim, S.: Dynamic Locomotion in the MIT Cheetah 3 Through Convex Model-Predictive Control. In: \emph{Proc. IEEE/RSJ Int. Conf. on Intelligent Robots and Systems (IROS)}, pp. 1--9. IEEE (2018). \doi{10.1109/IROS.2018.8594448}

\bibitem{wensing2023}
Wensing, P.M., Posa, M., Hu, Y., Escande, A., Mansard, N., Del Prete, A.: Optimization-Based Control for Dynamic Legged Robots. \emph{IEEE Transactions on Robotics} \textbf{40}, 43--63 (2024). \doi{10.1109/TRO.2023.3324580}

\bibitem{OCS2}
Farshidian, F., et al.: OCS2: An Open Source Library for Optimal Control of Switched Systems. Software available at: \url{https://github.com/leggedrobotics/ocs2} (accessed: 2026-11-16)

\bibitem{ProxQP}
Bambade, A., El-Kazdadi, S., Taylor, A., Carpentier, J.: PROX-QP: Yet another Quadratic Programming Solver for Robotics and beyond. In: \emph{Proc. Robotics: Science and Systems (RSS)} (2022). \doi{10.15607/RSS.2022.XVIII.040}

\bibitem{Eiquadprog}
Buondonno, G., Del Prete, A., Di Gaspero, L., Furfaro, A., Stephens, B., Guennebaud, G.: eiquadprog: C++ implementations of the Goldfarb--Idnani dual active-set method for convex quadratic programming. Software available at: \url{https://github.com/stack-of-tasks/eiquadprog} (accessed: 2026-11-16)

\bibitem{qpmad}
Sherikov, A.: qpmad: Eigen-based, header-only C++ implementation of the Goldfarb--Idnani dual active-set algorithm for quadratic programming. Software available at: \url{https://github.com/asherikov/qpmad} (accessed: 2026-11-16)

\bibitem{verschueren2018thesis}
Verschueren, R.: Convex Approximation Methods for Nonlinear Model Predictive Control. PhD thesis, Albert-Ludwigs-Universit\"at Freiburg im Breisgau (2018). \doi{10.6094/UNIFR/150231}

\bibitem{acados}
Verschueren, R., Frison, G., Kouzoupis, D., et al.: acados---a modular open-source framework for fast embedded optimal control. \emph{Mathematical Programming Computation} \textbf{14}(1), 147--183 (2022). \doi{10.1007/s12532-021-00208-8}

\bibitem{qpOASES}
Ferreau, H.J., Kirches, C., Potschka, A., Bock, H.G., Diehl, M.: qpOASES: A parametric active-set algorithm for quadratic programming. \emph{Mathematical Programming Computation} \textbf{6}(4), 327--363 (2014). \doi{10.1007/s12532-014-0071-1}

\bibitem{OSQP}
Stellato, B., Banjac, G., Goulart, P., Bemporad, A., Boyd, S.: OSQP: an operator splitting solver for quadratic programs. \emph{Mathematical Programming Computation} \textbf{12}(4), 637--672 (2020). \doi{10.1007/s12532-020-00179-2}

\bibitem{HPIPM}
Frison, G., Diehl, M.: HPIPM: a high-performance quadratic programming framework for model predictive control. \emph{IFAC-PapersOnLine} \textbf{53}(2), 6563--6569 (2020). \doi{10.1016/j.ifacol.2020.12.073}

\bibitem{qpSWIFT}
Pandala, A.G., Ding, Y., Park, H.-W.: qpSWIFT: A Real-Time Sparse Quadratic Program Solver for Robotic Applications.
\emph{IEEE Robotics and Automation Letters} \textbf{4}(4), 3355--3362 (2019). \doi{10.1109/LRA.2019.2926664}

\bibitem{SCS}
O'Donoghue, B., Chu, E., Parikh, N., Boyd, S.: Conic Optimization via Operator Splitting and Homogeneous Self-Dual Embedding. \emph{Journal of Optimization Theory and Applications} \textbf{169}(3), 1042--1068 (2016). \doi{10.1007/s10957-016-0892-3}. Software available at: \url{https://github.com/cvxgrp/scs} (accessed: 2026-11-16)

\bibitem{qpbenchmark}
Caron, S., Zaki, A., Otta, P., et al.: qpbenchmark: Benchmark for quadratic programming solvers available in Python. \url{https://github.com/qpsolvers/qpbenchmark} (accessed: 2026-11-16)

\bibitem{qpmad_benchmark}
Sherikov, A.: qpmad\_benchmark: Comparative benchmark of QP solvers (\texttt{qpmad}, \texttt{qpOASES}, \texttt{eiquadprog-fast}). \url{https://github.com/asherikov/qpmad_benchmark} (accessed: 2026-11-16)

\bibitem{daqp}
Arnstr\"om, D., Bemporad, A., Axehill, D.: A Dual Active-Set Solver for Embedded Quadratic Programming Using Recursive LDL$^\top$ Updates.
\emph{IEEE Transactions on Automatic Control} \textbf{67}(8), 4362--4369 (2022). \doi{10.1109/TAC.2022.3176430}

\bibitem{qpalm}
Hermans, B., Themelis, A., Patrinos, P.: QPALM: a proximal augmented Lagrangian method for nonconvex quadratic programs. \emph{Mathematical Programming Computation} \textbf{14}, 497--541 (2022). \doi{10.1007/s12532-022-00218-0}

\bibitem{piqp}
Schwan, R., Jiang, Y., Kuhn, D., Jones, C.N.: PIQP: A Proximal Interior-Point Quadratic Programming Solver. In: \emph{Proc. 62nd IEEE Conf. on Decision and Control (CDC)}, pp. 1088--1093. IEEE (2023). \doi{10.1109/CDC49753.2023.10383915}

\bibitem{cvxgen}
Mattingley, J., Boyd, S.: CVXGEN: A Code Generator for Embedded Convex Optimization.
\emph{Optimization and Engineering} \textbf{13}(1), 1--27 (2012). \doi{10.1007/s11081-011-9176-9}

\bibitem{cvxpygen}
Schaller, M., Banjac, G., Diamond, S., Agrawal, A., Stellato, B., Boyd, S.: Embedded Code Generation With CVXPY.
\emph{IEEE Control Systems Letters} \textbf{6}, 2653--2658 (2022). \doi{10.1109/LCSYS.2022.3173209}

\bibitem{forcespro}
Embotech AG: FORCESPRO Documentation (Solver Options / QP\_FAST, PDIP, ADMM).
Available at: \url{https://forces.embotech.com/documentation/solver_options/index.html} (accessed: 2026-11-16)

\bibitem{clarabel}
Goulart, P.J., Chen, Y.: Clarabel: An interior-point solver for conic programs with quadratic objectives. \emph{Mathematical Programming Computation} (2026). \doi{10.1007/s12532-026-00320-7}

\bibitem{ref_goldfarb1983}
Goldfarb, D., Idnani, A.: A Numerically Stable Dual Method for Solving Strictly Convex Quadratic Programs. \emph{Mathematical Programming} \textbf{27}(1), 1--33 (1983). \doi{10.1007/BF02591962}

\bibitem{Gurobi}
Gurobi Optimization, LLC: Gurobi Optimizer Reference Manual. Available at: \url{https://www.gurobi.com} (accessed: 2026-11-16)

\bibitem{MOSEK}
MOSEK ApS: The MOSEK Optimization Toolbox for Python Manual, Version 10.0. Available at: \url{https://www.mosek.com} (accessed: 2026-11-16)

\bibitem{qpmad_app}
Tuna, T., Nubert, J., Pfreundschuh, P., Cadena, C., Khattak, S., Hutter, M.: Informed, Constrained, Aligned: A Field Analysis on Degeneracy-Aware Point Cloud Registration in the Wild. \emph{IEEE Transactions on Field Robotics} \textbf{2}, 485--515 (2025). \doi{10.1109/TFR.2025.3576053}

\bibitem{schubiger2020gpu}
Schubiger, M., Banjac, G., Lygeros, J.: GPU acceleration of ADMM for large-scale quadratic programming. \emph{Journal of Parallel and Distributed Computing} \textbf{144}, 55--67 (2020). \doi{10.1016/j.jpdc.2020.05.021}

\bibitem{bishop2024reluqp}
Bishop, A.L., Zhang, J.Z., Gurumurthy, S., Tracy, K., Manchester, Z.: ReLU-QP: A GPU-Accelerated Quadratic Programming Solver for Model-Predictive Control. In: \emph{Proc. IEEE Int. Conf. on Robotics and Automation (ICRA)}, pp. 13285--13292. IEEE (2024). \doi{10.1109/ICRA57147.2024.10611249}

\bibitem{adabag2024mpcgpu}
Adabag, E., Atal, M., Gerard, W., Plancher, B.: MPCGPU: Real-Time Nonlinear Model Predictive Control through Preconditioned Conjugate Gradient on the GPU. In: \emph{Proc. IEEE Int. Conf. on Robotics and Automation (ICRA)}, pp. 9787--9794. IEEE (2024). \doi{10.1109/ICRA57147.2024.10611212}

\bibitem{amatucci2026mpx}
Amatucci, L., Sousa-Pinto, J., Turrisi, G., Orban, D., Barasuol, V., Semini, C.: Primal-Dual iLQR for GPU-Accelerated Learning and Control in Legged Robots. \emph{IEEE Robotics and Automation Letters} \textbf{11}(1), 1010--1017 (2026). \doi{10.1109/LRA.2025.3632610}

\bibitem{jeon2025cusadi}
Jeon, S.H., Hong, S., Lee, H.J., Khazoom, C., Kim, S.: CusADi: A GPU Parallelization Framework for Symbolic Expressions and Optimal Control. \emph{IEEE Robotics and Automation Letters} \textbf{10}(2), 899--906 (2025). \doi{10.1109/LRA.2024.3512254}

\bibitem{plancher2022grid}
Plancher, B., Neuman, S.M., Ghosal, R., Kuindersma, S., Janapa Reddi, V.: GRiD: GPU-Accelerated Rigid Body Dynamics with Analytical Gradients. In: \emph{Proc. IEEE Int. Conf. on Robotics and Automation (ICRA)}, pp. 6253--6260. IEEE (2022). \doi{10.1109/ICRA46639.2022.9812384}

\bibitem{fang2026gpusls}
Fang, J., Chou, G.: Safe Large-Scale Robust Nonlinear MPC in Milliseconds via Reachability-Constrained System Level Synthesis on the GPU. In: \emph{Proc. Robotics: Science and Systems (RSS)}, Sydney, Australia (2026). \doi{10.15607/RSS.2026.XXII.103}

\bibitem{alvarezpadilla2025wbmppi}
Alvarez-Padilla, J., Zhang, J.Z., Kwok, S., Dolan, J.M., Manchester, Z.: Real-Time Whole-Body Control of Legged Robots with Model-Predictive Path Integral Control. In: \emph{Proc. IEEE Int. Conf. on Robotics and Automation (ICRA)}, pp. 14721--14727. IEEE (2025). \doi{10.1109/ICRA55743.2025.11128271}

\bibitem{williams2017mppi}
Williams, G., Aldrich, A., Theodorou, E.A.: Model Predictive Path Integral Control: From Theory to Parallel Computation. \emph{Journal of Guidance, Control, and Dynamics} \textbf{40}(2), 344--357 (2017). \doi{10.2514/1.G001921}

\bibitem{xue2025dialmpc}
Xue, H., Pan, C., Yi, Z., Qu, G., Shi, G.: Full-Order Sampling-Based MPC for Torque-Level Locomotion Control via Diffusion-Style Annealing. In: \emph{Proc. IEEE Int. Conf. on Robotics and Automation (ICRA)}, pp. 4974--4981. IEEE (2025). \doi{10.1109/ICRA55743.2025.11127320}

\bibitem{katayama2023}
Katayama, S., Murooka, M., Tazaki, Y.: Model predictive control of legged and humanoid robots: models and algorithms. \emph{Advanced Robotics} \textbf{37}(5), 298--315 (2023). \doi{10.1080/01691864.2023.2168134}

\bibitem{kouzoupis2018}
Kouzoupis, D., Frison, G., Zanelli, A., Diehl, M.: Recent Advances in Quadratic Programming Algorithms for Nonlinear Model Predictive Control. \emph{Vietnam Journal of Mathematics} \textbf{46}, 863--882 (2018). \doi{10.1007/s10013-018-0311-1}

\bibitem{carpentier2021}
Carpentier, J., Wieber, P.-B.: Recent Progress in Legged Robots Locomotion Control. \emph{Current Robotics Reports} \textbf{2}, 231--238 (2021). \doi{10.1007/s43154-021-00059-0}

\bibitem{escande2014}
Escande, A., Mansard, N., Wieber, P.-B.: Hierarchical quadratic programming: Fast online humanoid-robot motion generation. \emph{International Journal of Robotics Research} \textbf{33}(7), 1006--1028 (2014). \doi{10.1177/0278364914521306}

\bibitem{kanoun2011}
Kanoun, O., Lamiraux, F., Wieber, P.-B.: Kinematic Control of Redundant Manipulators: Generalizing the Task-Priority Framework to Inequality Task. \emph{IEEE Transactions on Robotics} \textbf{27}(4), 785--792 (2011). \doi{10.1109/TRO.2011.2142450}

\bibitem{herzog2016}
Herzog, A., Rotella, N., Mason, S., Grimminger, F., Schaal, S., Righetti, L.: Momentum control with hierarchical inverse dynamics on a torque-controlled humanoid. \emph{Autonomous Robots} \textbf{40}, 473--491 (2016). \doi{10.1007/s10514-015-9476-6}

\bibitem{bellicoso2017}
Bellicoso, C.D., Jenelten, F., Fankhauser, P., Gehring, C., Hwangbo, J., Hutter, M.: Dynamic locomotion and whole-body control for quadrupedal robots. In: \emph{Proc. IEEE/RSJ Int. Conf. on Intelligent Robots and Systems (IROS)}, pp. 3359--3365. IEEE (2017). \doi{10.1109/IROS.2017.8206174}

\bibitem{sherikov2014}
Sherikov, A., Dimitrov, D., Wieber, P.-B.: Whole body motion controller with long-term balance constraints. In: \emph{Proc. IEEE-RAS Int. Conf. on Humanoid Robots}, pp. 444--450. IEEE (2014). \doi{10.1109/HUMANOIDS.2014.7041399}

\bibitem{maros1999}
Maros, I., M\'esz\'aros, C.: A repository of convex quadratic programming problems. \emph{Optimization Methods and Software} \textbf{11}(1--4), 671--681 (1999). \doi{10.1080/10556789908805768}

\bibitem{dolan2002}
Dolan, E.D., Mor\'e, J.J.: Benchmarking optimization software with performance profiles. \emph{Mathematical Programming} \textbf{91}, 201--213 (2002). \doi{10.1007/s101070100263}

\bibitem{ecos}
Domahidi, A., Chu, E., Boyd, S.: ECOS: An SOCP solver for embedded systems. In: \emph{Proc. European Control Conference (ECC)}, pp. 3071--3076. IEEE (2013). \doi{10.23919/ECC.2013.6669541}

\bibitem{highs}
Huangfu, Q., Hall, J.A.J.: Parallelizing the dual revised simplex method. \emph{Mathematical Programming Computation} \textbf{10}, 119--142 (2018). \doi{10.1007/s12532-017-0130-5}. The convex-QP capability of HiGHS is a later active-set addition: \url{https://github.com/ERGO-Code/HiGHS} (accessed: 2026-11-16)

\bibitem{cvxopt}
Andersen, M.S., Dahl, J., Vandenberghe, L.: CVXOPT: A Python package for convex optimization. Software available at: \url{https://cvxopt.org} (accessed: 2026-11-16)

\bibitem{tinympc}
Nguyen, K., Schoedel, S., Alavilli, A., Plancher, B., Manchester, Z.: TinyMPC: Model-Predictive Control on Resource-Constrained Microcontrollers. In: \emph{Proc. IEEE Int. Conf. on Robotics and Automation (ICRA)}. IEEE (2024). \doi{10.1109/ICRA57147.2024.10610987}

\bibitem{diehl2005}
Diehl, M., Bock, H.G., Schl\"oder, J.P.: A Real-Time Iteration Scheme for Nonlinear Optimization in Optimal Feedback Control. \emph{SIAM Journal on Control and Optimization} \textbf{43}(5), 1714--1736 (2005). \doi{10.1137/S0363012902400713}

\bibitem{gros2020}
Gros, S., Zanon, M., Quirynen, R., Bemporad, A., Diehl, M.: From linear to nonlinear MPC: bridging the gap via the real-time iteration. \emph{International Journal of Control} \textbf{93}(1), 62--80 (2020). \doi{10.1080/00207179.2016.1222553}

\bibitem{cimini2017}
Cimini, G., Bemporad, A.: Exact Complexity Certification of Active-Set Methods for Quadratic Programming. \emph{IEEE Transactions on Automatic Control} \textbf{62}(12), 6094--6109 (2017). \doi{10.1109/TAC.2017.2696742}

\bibitem{arnstrom2022}
Arnstr\"om, D., Axehill, D.: A Unifying Complexity Certification Framework for Active-Set Methods for Convex Quadratic Programming. \emph{IEEE Transactions on Automatic Control} \textbf{67}(6), 2758--2770 (2022). \doi{10.1109/TAC.2021.3090749}

\bibitem{bemporad2002}
Bemporad, A., Morari, M., Dua, V., Pistikopoulos, E.N.: The explicit linear quadratic regulator for constrained systems. \emph{Automatica} \textbf{38}(1), 3--20 (2002). \doi{10.1016/S0005-1098(01)00174-1}

\bibitem{posa2014}
Posa, M., Cantu, C., Tedrake, R.: A direct method for trajectory optimization of rigid bodies through contact. \emph{International Journal of Robotics Research} \textbf{33}(1), 69--81 (2014). \doi{10.1177/0278364913506757}

\bibitem{lecleach2024}
Le Cleac'h, S., Howell, T.A., Yang, S., Lee, C.-Y., Zhang, J., Bishop, A., Schwager, M., Manchester, Z.: Fast Contact-Implicit Model Predictive Control. \emph{IEEE Transactions on Robotics} \textbf{40}, 1617--1629 (2024). \doi{10.1109/TRO.2024.3351554}

\bibitem{stewart1996}
Stewart, D.E., Trinkle, J.C.: An implicit time-stepping scheme for rigid body dynamics with inelastic collisions and Coulomb friction. \emph{International Journal for Numerical Methods in Engineering} \textbf{39}(15), 2673--2691 (1996)

\bibitem{grandia2023}
Grandia, R., Jenelten, F., Yang, S., Farshidian, F., Hutter, M.: Perceptive Locomotion Through Nonlinear Model-Predictive Control. \emph{IEEE Transactions on Robotics} \textbf{39}(5), 3402--3421 (2023). \doi{10.1109/TRO.2023.3275384}

\bibitem{kim2019}
Kim, D., Di Carlo, J., Katz, B., Bledt, G., Kim, S.: Highly Dynamic Quadruped Locomotion via Whole-Body Impulse Control and Model Predictive Control. \emph{arXiv preprint} arXiv:1909.06586 (2019)

\bibitem{zeilinger2014}
Zeilinger, M.N., Morari, M., Jones, C.N.: Soft Constrained Model Predictive Control With Robust Stability Guarantees. \emph{IEEE Transactions on Automatic Control} \textbf{59}(5), 1190--1202 (2014). \doi{10.1109/TAC.2014.2304371}

\bibitem{nasoq}
Cheshmi, K., Kaufman, D.M., Kamil, S., Dehnavi, M.M.: NASOQ: numerically accurate sparsity-oriented QP solver. \emph{ACM Transactions on Graphics} \textbf{39}(4) (2020). \doi{10.1145/3386569.3392486}

\bibitem{proxqp2025}
Bambade, A., Schramm, F., El-Kazdadi, S., Caron, S., Taylor, A., Carpentier, J.: PROXQP: an Efficient and Versatile Quadratic Programming Solver for Real-Time Robotics Applications and Beyond. \emph{IEEE Transactions on Robotics} (2025). \doi{10.1109/TRO.2025.3577107}

\bibitem{cupiqp}
PREDICT-EPFL: cuPIQP --- batched and differentiable proximal interior-point QP solver on GPU. Software available at: \url{https://github.com/PREDICT-EPFL/cupiqp} (accessed: 2026-11-16)

\end{thebibliography}
